\documentclass{article}
\usepackage{log_2026}						

\usepackage{booktabs}						
\usepackage{multirow}						
\usepackage{amsfonts}						
\usepackage{graphicx}						
\usepackage{duckuments}						

\usepackage{booktabs}						
\usepackage{multirow}						
\usepackage{amsfonts}						
\usepackage{graphicx}						
\usepackage{duckuments}						
\usepackage[numbers,compress,sort]{natbib}	
\usepackage[utf8]{inputenc} 
\usepackage[T1]{fontenc}    
\usepackage{url}            
\usepackage{booktabs}       
\usepackage{amsfonts}       
\usepackage{nicefrac}       
\usepackage{microtype}      
\usepackage{xcolor}         

\usepackage{multicol}
\usepackage{amsmath}
\usepackage{subcaption}
\usepackage{enumitem}

\newcommand{\R}{\mathbb{R}}

\newcommand{\ii}{\mathrm{i}}
\newcommand{\wt}[1]{\widetilde{#1}}
\newcommand{\vect}[1]{\mathbf{#1}}
\newcommand{\set}[1]{\mathcal{#1}}
\newcommand{\norm}[1]{\left\lVert #1 \right\rVert}
\newcommand{\abs}[1]{\left\lvert #1 \right\rvert}
\newcommand{\concat}{\mathbin\Vert}
\newcommand\blfootnote[1]{%
  \begingroup
  \renewcommand\thefootnote{}\footnote{#1}%
  \addtocounter{footnote}{-1}%
  \endgroup
}
\usepackage[numbers,compress,sort]{natbib}	

\title[Signed Graph Pre-Training and Prompt Learning]{Signed Graph Pre-Training and Prompt Learning}

\author[Mei et al.]{%
Zihan Mei\\
Arizona State University\\
\email{zihanmei@asu.edu}\And
Rong Pan\\
Arizona State University\\
\email{Rong.Pan@asu.edu}\And
Yuzhou Chen\\
University of California, Riverside\\
\email{Yuzhou.Chen@ucr.edu}\And
Yixuan He\thanks{Corresponding author.}\\
Arizona State University\\
\email{Yixuan.He@asu.edu}
}

\begin{document}

\maketitle

\begin{abstract}
Signed graphs arise in trust--distrust networks, financial correlation systems, biological interaction graphs, and many other domains in which edges can be positive or negative and may also be directed. While signed graph neural networks have improved task-specific learning, graph transfer learning on signed graphs remains underdeveloped. In this paper, we introduce TopoSIGN, a pioneering topology-guided graph pre-training and prompt learning framework for signed graphs. TopoSIGN combines a structural encoder built on the magnetic signed Laplacian with a novel persistent-homology branch that summarizes signed topology through Dowker-complex persistence images. The fused embeddings are then transferred to a prompt learning function. Experimental results on synthetic and real-world datasets demonstrate the efficacy of TopoSIGN in extracting useful structural information in signed graphs, as well as the adaptability and flexibility of the proposed general framework.
\end{abstract}

\section{Introduction}
\label{sec:intro}
Graphs with both positive and negative interactions appear naturally in applications where agreement and disagreement, promotion and inhibition, or attraction and repulsion coexist~\cite{leskovec2010predicting,he2022sssnet, he2024pytorch}, including trust and distrust links in online social networks, competitive and synergistic relations in biological systems, and positively or negatively correlated financial assets. In many of these applications, edge polarity is as informative as connectivity itself, and in some cases, the graph is also directed~\cite{he2022msgnn}, so both sign and orientation must be modeled jointly. These properties make signed graph representation learning fundamentally different from the standard unsigned setting.

Recent work on signed graph neural networks (GNNs) has made substantial progress on link sign prediction, node classification, and node clustering by explicitly modeling positive and negative relations, balance theory, or signed random walks~\cite{he2022sssnet,he2022msgnn, derr2018signed,huang2021sdgnn,  fiorini2023sigmanet, zhao2025robust}. In particular, MSGNN~\cite{he2022msgnn} extends the magnetic Laplacian from directed graphs to a magnetic signed Laplacian, thereby handling directed signed graphs within one Hermitian spectral framework. However, most signed GNNs remain largely task-specific and are trained end-to-end for a single supervised objective. By contrast, graph pre-training, prompt learning, and graph foundation models aim to learn transferable representations that can be adapted to multiple downstream tasks with limited labels~\cite{sun2022gppt, sun2023all}. This line of work has been highly active on unsigned graphs, but signed graphs have received far less attention. SGPT~\cite{zhai2025sgpt} is the only existing graph pre-training and fine-tuning work on signed graphs so far, but its goal is fundamentally different from ours: SGPT transfers knowledge from unsigned graph learning to signed learning tasks, while here we pretrain and fine-tune on signed tasks directly.

Topology is another underused source of signal in signed graph representation learning. Persistent homology (PH) and related topological data analysis (TDA)~\cite{wasserman2018topological,chazal2021introduction,carlsson2012topological,chen2025topological,dixon2025topological,chen2021zigzag} tools provide multi-scale summaries of higher-order connectivity patterns beyond local neighborhoods. The recent directed-graph framework TopoDIG~\cite{liang2026topologyguided} combined a magnetic Laplacian encoder with a Dowker-complex topological branch~\cite{dowker1952topology} for pre-training and prompting on directed unsigned graphs. However, extending that idea to signed graphs is not trivial. In addition to a sign-aware GNN encoder, the TDA branch itself must be reconsidered: negative edges do not fit naturally into the unsigned filtration design used in prior work, and naively injecting signed weights into a scalar filtration can produce a poor match between the filtration semantics and the intended persistent-homology interpretation.

In this paper, we design for signed graph pre-training a two-branch structure, i.e., a spectral graph encoder plus a Dowker complex-based topological encoder, ensuring that they are aligned with the semantic needs of signed graphs. Concretely, we propose \emph{TopoSIGN}, a \textbf{Topo}logy-guided pre-training and prompt learning model for \textbf{sign}ed graphs. First, we employ a signed directed GNN method, e.g., SSSNET~\cite{he2022sssnet} or MSGNN~\cite{he2022msgnn}, for the structural encoder. 
Second, and more importantly, we introduce a signed degree-vector distance filtration that does not neglect signed information. That is, instead of assigning each node a single scalar degree in the filtration definition, we associate with each node a two-dimensional vector containing its positive and negative degrees and define a novel Dowker filtration through distances in this signed degree space. This allows us to retain signed information inside the simple yet effective topological branch without resorting to multiparameter persistence. After that, we fuse the learned embeddings of the two branches and apply a graph prompt learning paradigm~\cite{zi2024prog,sun2022gppt} for graph pre-training and fine-tuning. 

The main contributions of this work are as follows.

\begin{itemize}[noitemsep, leftmargin=*]
    \item We formulate \emph{TopoSIGN}, the first topology-guided graph pre-training and prompt-learning framework for signed graphs.
    \item We design for signed graphs a topology-empowered graph prompt function that improves the transfer and generalization capabilities of GNNs.
    \item We conduct experiments on both synthetic and real-world datasets with two instantiations of our framework and demonstrate its efficacy and flexibility. The framework is general and compatible with various signed GNNs and prompt learning functions.
\end{itemize}

\blfootnote{Code is available at \url{https://github.com/Harrison-zh-M/TopoSIGN}.}
\section{Related work}
\label{sec:related_work}
\subsection{Signed graph representation learning}
Signed graph learning has a long history in spectral clustering, matrix factorization, and balance-theoretic embedding~\cite{cucuringu2019sponge,rafailidis2016modeling,zhang2024signed}. Neural approaches extend these ideas by designing message passing or spectral filters that distinguish positive and negative relations. SGCN uses balance theory to propagate separate representations through positive and negative neighborhoods~\cite{derr2018signed}. SiGAT and SDGNN incorporate signed directed motifs and signed aggregation objectives for link sign prediction on directed signed networks~\cite{huang2019signed,huang2021sdgnn}. SNEA learns signed network embeddings with graph attention~\cite{li2020learning}. For clustering, SSSNET introduces a semi-supervised signed clustering objective that is not restricted to strong balance assumptions~\cite{he2022sssnet}. Spectral signed GNNs include SLGNN~\cite{li2023signed}, frequency-based signed GNNs~\cite{chen2024graph}, SigMaNet~\cite{fiorini2023sigmanet}, and MSGNN~\cite{he2022msgnn}, which define signed or signed-directed Laplacian operators for signed representation learning. DSGC studies robust deep signed graph clustering through weak balance theory and graph denoising~\cite{zhao2025robust}. SE-SGformer develops a self-explainable signed graph transformer for link sign prediction with signed random-walk positional encodings~\cite{li2025self}. CopulaLSP models inter-edge dependencies for scalable link sign prediction on signed graphs~\cite{sung2026scalable}. Here we pick MSGNN~\cite{he2022msgnn} as one of our structural backbones since it is simple and effective to complement the more computationally expensive topological branch for our task, and it can directly handle the tasks of our interest, i.e., link prediction and node clustering. SSSNET~\cite{he2022sssnet} is also picked due to its simplicity and efficacy.
\subsection{Graph pre-training, prompting, and graph foundation models}
Graph pre-training learns reusable representations through self-supervised objectives at node, edge, or graph level. Representative methods include contrastive approaches such as DGI~\cite{velickovic2018deep} and GraphCL~\cite{you2020graph}, masked or reconstruction-based models such as GraphMAE~\cite{hou2022graphmae}
and GraphMAE2~\cite{hou2023graphmae2}, and task-specific pre-training frameworks such as S2PGNN~\cite{zhili2024search} and BRep-BERT~\cite{lou2023brep}. Prompt learning attempts to reduce the gap between pre-training and downstream tasks. GPPT uses edge prediction as the pretext task and introduces tokens to align node classification with pre-training~\cite{sun2022gppt}; GraphPrompt unifies graph pre-training and downstream tasks with prompt-based readout~\cite{liu2023graphprompt}; GPF learns universal graph prompts~\cite{fang2023universal}; and All-in-One formulates multi-task graph prompting through meta-learning~\cite{sun2023all}. ProG provides a benchmark and systematic evaluation for graph prompt learning~\cite{zi2024prog}, while GCOPE~\cite{zhao2024all} and OpenGraph~\cite{xia2024opengraph} study cross-domain transfer and open graph foundation modeling, respectively. SAMGPT focuses on text-free multi-domain graph pre-training and cross-domain adaptation~\cite{yu2025samgpt}. UniGraph learns a unified cross-domain model for text-attributed graphs~\cite{he2025unigraph}. GraphTOP adapts pre-trained GNNs by topology-oriented prompting through local edge rewiring~\cite{fu2025graphtop}. TopoDIG~\cite{liang2026topologyguided} fuses topological features and a directed graph neural network encoder into the GPPT-like framework and extends the pre-training and prompt learning framework to directed graphs with asymmetric sending and receiving patterns. These methods establish strong transfer baselines, but most assume unsigned or text-attributed graphs and do not natively model signed Laplacian structure or signed topological filtrations. SGPT~\cite{zhai2025sgpt} is the only existing graph pre-training and fine-tuning work on signed graphs so far, but their goal is fundamentally different from ours: SGPT transfers knowledge from unsigned graph learning to signed learning tasks, while here we pretrain and fine-tune on signed tasks directly. Our approach gets inspiration from TopoDIG~\cite{liang2026topologyguided} by constructing a signed directed graph encoder and a topological encoder and feeding the concatenated outputs into a GPPT-like prompting framework.
\subsection{Topological learning on graphs}

Recent TDA-based graph representation learning uses persistent homology and related topological summaries to encode connectivity, cycles, and higher-order structures that are difficult for standard message passing to capture. Early neural and kernel-based methods include PersLay, which learns vectorized persistence-diagram representations and introduces graph topological signatures~\cite{carriere2020perslay}, P-WL, which augments Weisfeiler--Lehman subtree features with persistent cycle information~\cite{rieck2019persistent}, and Graph Filtration Learning, which learns a differentiable filtration function and uses persistent homology as a graph-level readout~\cite{hofer2020graph}. Later methods integrate topology more directly into GNN architectures: TRI-GNN rewires local graph neighborhoods using persistent homology and uses topological summaries as side information for node classification~\cite{chen2021topological}; TOGL injects global persistent-homology features into message-passing GNNs and improves expressiveness beyond standard WL-limited aggregation~\cite{horn2022topological}; and Wit-TopoPool uses persistent homology and witness complexes to design topology-aware pooling for graph classification~\cite{chen2023topological}. More recent work shifts from graph-level summaries to localized and task-adaptive topological features. TTG-NN combines persistent homology, graph convolution, and tensor operations to capture local and global graph structure~\cite{wen2024tensor}; Yan et al.~\cite{yan2025enhancing} use extended persistent homology on vicinity graphs to construct node- and edge-level topological features for node classification and link prediction; TensorMV-GCL incorporates extended persistent homology into multi-view graph contrastive learning~\cite{wu2024tensor}; and GraphTCL aligns GNN structural embeddings with persistent-homology embeddings through cross-view contrastive learning~\cite{korkmaz2025cross}. TopoDIG extends this direction to directed graphs by combining magnetic-Laplacian-based directed graph encoding with Dowker-complex-based topological features in a pre-training and prompting framework~\cite{liang2026topologyguided}. Despite this progress, almost all TDA-enhanced graph learning methods assume unsigned or merely weighted graphs. For signed graphs, existing topological work mainly characterizes structural balance using simplicial homology and cohomology~\cite{she2019algebraic}, but does not provide a modern pre-training, prompting, or GNN representation-learning framework. Thus, signed topological graph learning remains largely open. 

\section{Methodology}
\label{sec:method}
\begin{figure}[htb]
\centering
\includegraphics[width=\textwidth]{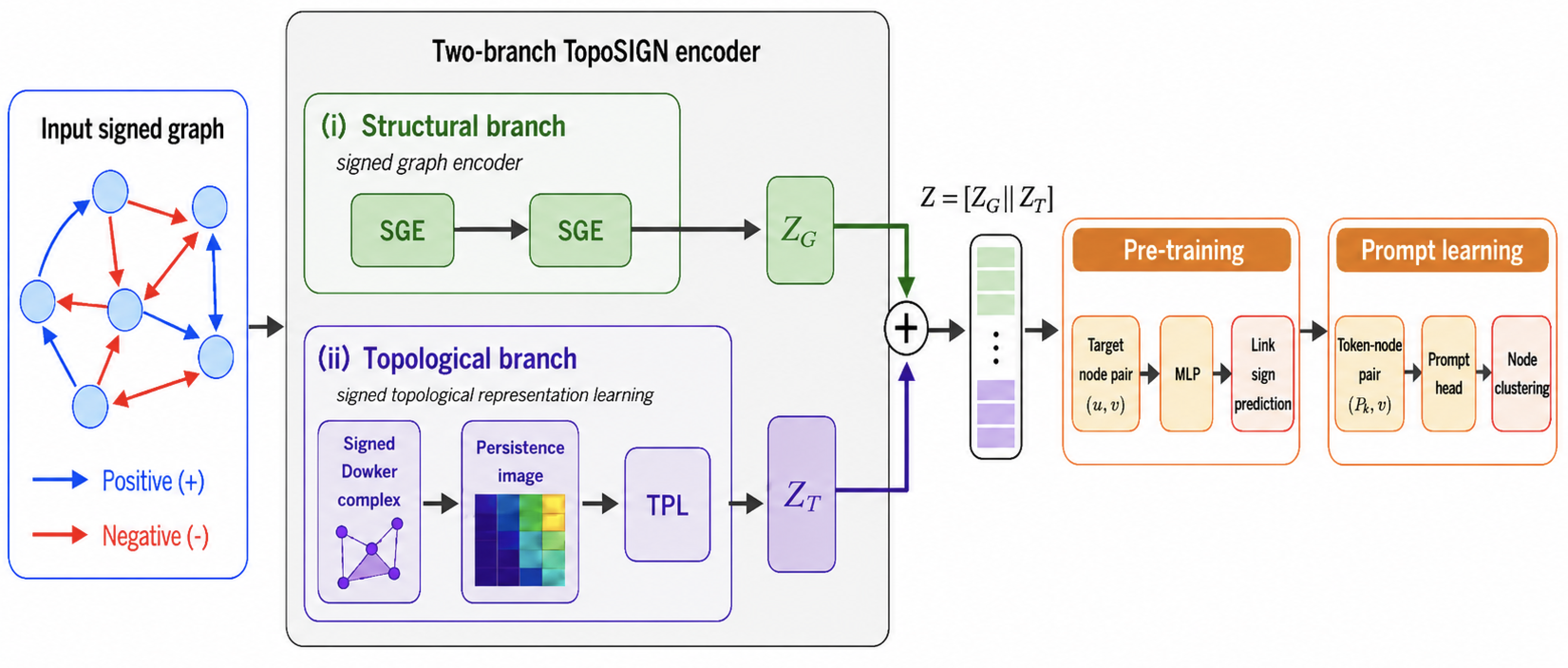}
\caption{\textcolor{black}{Overview of the TopoSIGN pre-training and prompting framework. Given an input signed (possibly also directed) graph with node attributes, TopoSIGN learns transferable representations through two complementary pathways (with the gray box). The first branch (i) applies signed graph encoder (SGE) layers to capture signed, directional, and local structural information. In parallel, the second branch (ii) constructs a Dowker complex to encode higher-order relational structures, which are transformed into compact topological embeddings via the topological representation learning (TPL) module using a projection layer. The structural and topological representations are fused to form joint node embeddings. Pre-training is performed using a pairwise prediction objective on both node-node and token-node pairs with a shared MLP and pretext loss, enabling topology-aware and prompt-compatible representations for downstream tasks.}\label{fig:toposign_overview}}
\end{figure}
\subsection{Problem formulation}
\label{sec:formulation}
Let $\set{G} = (\set{V}, \set{E}, w, \vect{X})$
be a signed graph, possibly directed and weighted, where $\set{V}=\{v_1,\ldots,v_n\}$ is the node set, $\set{E}\subseteq \set{V}\times\set{V}$ is the edge set, $w: \set{E} \rightarrow \R\setminus\{0\}$ assigns a nonzero signed weight to each edge, and $\vect{X}\in\R^{n\times F}$ is the node-feature matrix. We denote positive and negative edge sets by $\set{E}^{+}=\{(u,v)\in\set{E}: w(u,v)>0\}$ and
$\set{E}^{-}=\{(u,v)\in\set{E}: w(u,v)<0\}.$
The signed adjacency matrix $\vect{A}\in\R^{n\times n}$ is defined entrywise as $A_{uv}=w(u,v)$ if $(u,v)\in\set{E}$ and zero otherwise.
We further define the positive and negative magnitude matrices
$\vect{A}^{+}=\max(\vect{A},0)$ and $\vect{A}^{-}=\max(-\vect{A},0)$, where the maximum is taken elementwise.

The pre-training task does not involve any node labels and is based completely on observed edges. Following \cite{he2022msgnn}, we consider different variations of the link prediction task for \emph{signed and/or directed} networks. In our default pre-training task, link sign prediction (SP), one assumes that there is a link from $u$ to $v$ and aims to predict whether that link is positive or negative, i.e., whether $(u, v) \in \mathcal{E}^+$ or $(u, v) \in \mathcal{E}^-$. 
The downstream task is node clustering, whose goal is to partition the node set into a disjoint union of $K$ clusters,
$\set{V}=\set{C}_0 \cup \set{C}_1 \cup \cdots \cup \set{C}_{K-1}$.

{\bf Persistent Homology}
Persistent homology (PH) provides a principled framework for analyzing filtered simplicial complexes~\cite{edelsbrunner2008persistent,zomorodian2004computing,otter2017roadmap}. Early applications primarily focused on point clouds $\mathcal{X} \subset \mathbb{R}^N$, where Vietoris-Rips filtrations construct a sequence of nested complexes $\Delta_1(\mathcal{X}) \subset \Delta_2(\mathcal{X}) \subset \cdots$, enabling the tracking of topological features across multiple scales~\cite{de2022ripsnet,turkes2022effectiveness}. The resulting persistence diagram
$\mathcal{D}_q(\mathcal{X}) = \{(b_i, d_i)\}$
records the birth and death times of $q$-dimensional features
(where $q \in \{0, 1, \dots\}$), with longer lifetimes
$(d_i - b_i)$ typically interpreted as more structurally significant. PH has also been extended beyond point clouds to structured domains such as graphs and images. Unlike Euclidean data, graphs lack an intrinsic notion of scale, making the choice of filtration nontrivial. To address this, two principal approaches have been proposed. The first is the power-based filtration, which treats nodes as points in a metric space induced by graph distances and constructs a Vietoris-Rips filtration~\cite{sheehy2012linear,aktas2019persistence}. While conceptually straightforward, this approach is often computationally expensive. The second, more practical approach is based on {sublevel set filtrations}, where a scalar function $f$ defined on nodes or edges induces a sequence of nested subgraphs~\cite{hofer2017deep,hofer2020graph}. These subgraphs are then lifted to simplicial complexes via clique expansions. A key interpretability distinction arises between these approaches. In power filtrations, persistence reflects geometric scale, whereas in sublevel set filtrations, it reflects variations in the function $f$. Consequently, long persistence does not necessarily correspond to large-scale structures. The standard PH pipeline for graphs consists of three steps, i.e., {filtration}, {persistence computation}, and {vectorization}. Given a graph $\mathcal{G} = (\mathcal{V}, \mathcal{E})$ and a function $f : \mathcal{G} \to \mathbb{R}$, a sequence of thresholds $\{\alpha_i\}_{i=1}^n$ induces nested subgraphs $\mathcal{G}_1 \subset \cdots \subset \mathcal{G}_n$, $\mathcal{G}_i \subset \mathcal{G}$ satisfies $f(\mathcal{G}_i) \le \alpha_i$. Each subgraph is lifted to its clique complex $\hat{\mathcal{G}}_i$, forming a filtration $\{\hat{\mathcal{G}}_i\}$. Persistence diagrams $\mathcal{D}_q(\mathcal{G}, f) = \{(b_j, d_j)\}$ summarize the birth and death of homological features $H_q(\hat{\mathcal{G}}_i)$, and can be converted into fixed-length representations via persistence images or persistence landscapes~\cite{adams2017persistence,bubenik2015statistical}.

Despite their effectiveness, these constructions often suffer from significant computational overhead, particularly for large graphs due to the combinatorial growth of simplicial complexes. To mitigate this challenge, in this paper, we adopt the {Dowker complex}~\cite{dowker1952topology}, which constructs simplicial structures based on a bipartite relation between a subset of representative nodes (landmarks) and the remaining nodes (witnesses). That is, given a graph $\mathcal{G}$, a landmark set $L \subseteq \mathcal{V}$ and a witness set $W = \mathcal{V} \setminus L$ are defined, and simplices are formed based on proximity relations between landmarks and witnesses. This construction preserves essential topological features while significantly reducing computational complexity. 
\subsection{Overall framework of TopoSIGN}
\label{subsec:overview}
TopoSIGN consists of three modules: a structural signed GNN encoder (SGE, we apply MSGNN~\cite{he2022msgnn} and SSSNET~\cite{he2022sssnet} here), a signed topological encoder (with our novel design adapted from \cite{liang2026topologyguided}), and a prompt module (we apply similar architectures as in \cite{sun2022gppt}). The structural encoder maps the signed graph to node embeddings $\vect{Z}_{\set{G}}\in\R^{n\times d_G}$. The topological encoder computes persistence-image features from a Dowker filtration and maps them to $\vect{Z}_{\set{T}}\in\R^{n\times d_T}$. The fused node representation is $\vect{Z}=\big[\vect{Z}_{\set{G}}\,\concat\,\vect{Z}_{\set{T}}\big]\in\R^{n\times(d_G+d_T)},$
where $[\cdot\concat\cdot]$ denotes the concatenation function. The same fused representation is used for (link sign) pre-training and (cluster) prompt adaptation. A framework overview is provided in Fig.~\ref{fig:toposign_overview}.

\subsection{Signed graph encoder (SGE)}
The SGE branch is responsible for modeling graph structural information in terms of sign, direction, and spectral smoothness over the full signed graph, and may contain one or more SGE layers using different SGE backbones from signed GNNs.

One instantiation of the structural branch of TopoSIGN uses the magnetic signed Laplacian introduced by \cite{he2022msgnn} to encode signed local and global structure. First, define the symmetrized signed adjacency and degree matrices by
\begin{equation}
    \wt{\vect{A}}_{uv}=\frac{1}{2}(A_{uv}+A_{vu}),
    \qquad
    \wt{\vect{D}}_{uu}=\frac{1}{2}\sum_{v=1}^{n}\big(\abs{A_{uv}}+\abs{A_{vu}}\big).
\end{equation}
The directional asymmetry is encoded through the phase matrix
$\vect{\Theta}^{(q)}_{uv}=2\pi q\,(A_{uv}-A_{vu}),$
where $q\in\R$ is the ``charge parameter''. The corresponding Hermitian adjacency matrix is $\vect{H}^{(q)} = \wt{\vect{A}}\odot \exp(\ii\vect{\Theta}^{(q)}),$
where $\odot$ denotes the Hadamard product. MSGNN then defines the unnormalized and normalized magnetic signed Laplacian matrices by
\begin{equation}
    \vect{L}_{U}^{(q)} = \wt{\vect{D}} - \vect{H}^{(q)},
    \qquad
    \vect{L}_{N}^{(q)} = \vect{I} - \left(\wt{\vect{D}}^{-1/2}\wt{\vect{A}}\wt{\vect{D}}^{-1/2}\right)\odot \exp(\ii\vect{\Theta}^{(q)}).
\end{equation}
\cite{he2022msgnn} defines spectral convolution by diagonalizing a Hermitian Laplacian $\vect{L}$ and approximating the resulting filter with Chebyshev polynomials. In our instantiation, with
$\vect{L}=\vect{L}_{N}^{(q)}$ and choosing the second-order polynomial approximation, the layer update becomes
\begin{equation}
\vect{X}^{(\ell)} =
\sigma\!\left(
\vect{X}^{(\ell-1)}\vect{W}_{\mathrm{self}}^{(\ell)}
+ \wt{\vect{L}}_{N}^{(q)}\vect{X}^{(\ell-1)}\vect{W}_{\mathrm{neigh}}^{(\ell)}
+ \vect{B}^{(\ell)}
\right),
\end{equation}
where $\mathbf{W}_{\text{self}}^{(\ell)}$ and $\mathbf{W}_{\text{neigh}}^{(\ell)}$ are learned weight matrices corresponding to the filter weights of different channels and $\mathbf{B}^{(\ell)}$ is the bias vector, $\wt{\vect{L}}_{N}^{(q)}=\frac{2}{\lambda_{\max}} \vect{L}_{N}^{(q)}-\vect{I}$ is the rescaled Laplacian with identity matrix $\vect{I}$, and $\sigma$ is a nonlinear activation function. As in \cite{zhang2021magnet, he2022msgnn}, we use a complex version of the Rectified Linear Unit defined by $\sigma(z)=z$, if $-\pi/2\leq \arg(z)<\pi/2$, and $\sigma(z)=0$ otherwise, where $\arg(\cdot)$ is the complex argument of $z\in\mathbb{C}$. After the final layer, the complex
representation is unwound into a real-valued embedding by concatenating its real and imaginary
parts. We use $q=0.25$ by default. 

Another instantiation we use is SSSNET (SIMPA) from \cite{he2022sssnet}. Its core component is the SIMPA aggregation mechanism, which separately propagates information over positive and negative neighborhoods to preserve the distinct semantics of friendship and antagonism relations. Specifically, SSSNET computes decoupled embeddings from multiple signed propagation paths and combines them to capture agreement, disagreement, and neutral interactions across the graph. Owing to its lightweight propagation structure and effectiveness on signed community detection benchmarks, SSSNET serves as a simple yet strong structural encoder in our framework.

\subsection{Signed topological representation learning}
The topological branch complements the SGE embedding with a multiscale signed summary derived from persistent homology on Dowker complexes. The key design choice is how to define a scalar filtration that remains meaningful for signed graphs.

\paragraph{Positive-subgraph baseline.}
The simplest sign-compatible baseline removes negative edges before topology is computed. Let
$\set{G}^{+}=(\set{V},\set{E}^{+},\vect{A}^{+})$
be the positive subgraph. Its degree for node $u$ is $d^{+}(u)=\sum_{v=1}^{n}\big(A^{+}_{uv}+A^{+}_{vu}\big).$

Given landmark and witness sets $L,W\subseteq \set{V}$, we define a scalar landmark--witness distance by
\begin{equation}
    \delta_{+}(\ell,w)=\abs{d^{+}(\ell)-d^{+}(w)}.
\end{equation}
For threshold $\varepsilon\ge 0$, the associated Dowker complex is
\begin{equation}
    D_{\varepsilon}^{+}(L,W)=\left\{\sigma\subseteq L\;\middle|\; \exists w\in W\text{ such that }\delta_{+}(\ell,w)\le \varepsilon,\;\forall\ell\in\sigma\right\}.
\end{equation}
Sweeping $\varepsilon$ produces a one-parameter filtration. This baseline is useful because it keeps the PH semantics clean and provides a direct comparison point for any sign-aware topological design.

\paragraph{Signed degree-vector distance filtration.}
Our proposed topological variant uses the full signed graph instead of discarding negative edges. For each node $u$, define the positive and negative degree magnitudes
\begin{equation}
    d^{+}(u)=\sum_{v=1}^{n}\big(A^{+}_{uv}+A^{+}_{vu}\big),
    \qquad
    d^{-}(u)=\sum_{v=1}^{n}\big(A^{-}_{uv}+A^{-}_{vu}\big).
\end{equation}
The negative degree is nonnegative by construction because $\vect{A}^{-}$ stores absolute magnitudes of negative edges. We then associate to node $u$ the two-dimensional signed degree vector
\begin{equation}
    \vect{s}(u)=
    \begin{bmatrix}
        d^{+}(u) \\
        d^{-}(u)
    \end{bmatrix}\in\R^{2}.
\end{equation}
The landmark--witness distance is defined by Euclidean distance in this degree space
\begin{equation}
    \delta_{\pm}(\ell,w)=\norm{\vect{s}(\ell)-\vect{s}(w)}_2.
    \label{eq:vecdist}
\end{equation}
The resulting signed Dowker filtration is
\begin{equation}
    D_{\varepsilon}^{\pm}(L,W)=\left\{\sigma\subseteq L\;\middle|\; \exists w\in W\text{ such that }\delta_{\pm}(\ell,w)\le \varepsilon,\;\forall\ell\in\sigma\right\}.
\end{equation}
This construction keeps the filtration one-dimensional, which allows standard PH software and persistence-image vectorizations to be used unchanged, while preserving information about how much positive and negative connectivity each node carries. Nodes with similar total degree but very different sign composition are therefore distinguished by the filtration. Additionally, to further improve scalability, we adopt a landmark–witness strategy: a subset of nodes is selected as landmarks (e.g., via either degree- or centrality-based sampling), which serve as the backbone for the filtration, while the remaining nodes act as witnesses that attach to nearby landmarks according to their signed connectivity. This design preserves the essential signed structural patterns while reducing computational complexity.

\paragraph{Persistence images and topological encoder.}
From a filtration family we compute persistence diagrams in dimensions $0$ and $1$ and then convert them to persistence images (PIs) \citep{adams2017persistence}. Since here we are interested in node-level representations, for each node, we construct topological features based on its two-hop ego-network (much smaller than the whole network). For each two-hop ego-network, we define a filtration based on node degrees. This yields a nested sequence of subgraphs, from which we compute \(H_{0}\) (connectivity) and \(H_{1}\) (cycles) persistence diagrams that capture the local topology across scales. Because these diagrams vary in size across nodes, we transform them into PIs—a fixed-dimensional vector representation. We achieve this by mapping the diagrams to {\it (birth, persistence)} coordinates, convolving the points with a Gaussian kernel weighted by a persistence-dependent function, and discretizing the resulting surface into a \(d_I \times d_I\) grid, where $d_I$ denotes the resolution of the image. This ensures a consistent input size for downstream machine learning models while preserving the topological signatures of the local neighborhood. Flattening the grid into a vector of a fixed length $d_T=d_I^2$ gives the topological representations of each node.

Let $\vect{Z}_{\set{T}}\in \R^{n\times d_T}$ contain all topological feature vectors in its rows, we then fuse it with the SGE embedding:
\begin{equation}
\vect{Z}=\big[\vect{Z}_{\set{G}}\,\concat\,\vect{Z}_{\set{T}}\big]\in\R^{n\times(d_G+d_T)}.
    \label{eq:fusion}
\end{equation}
The fused representation $\vect{Z}$ is used by both the pre-training loss and the downstream prompt head.

\subsection{Pre-training with link sign prediction}
The pre-training task is link sign prediction. Let $\Omega_{\mathrm{tr}}\subseteq\set{E}$ denote the training edge set produced by a signed edge split. Each training edge receives a binary sign label,
\begin{equation}
    y_{uv}=\begin{cases}
        1, & A_{uv}>0,\\
        0, & A_{uv}<0.
    \end{cases}
\end{equation}
Given the fused node embeddings $\vect{z}_u$ and $\vect{z}_v$ from Eq.~\eqref{eq:fusion}, an edge decoder $\psi_{\zeta}(\cdot)$ predicts the edge sign,
\begin{equation}
    \hat{y}_{uv}=\operatorname{sigmoid}(\cdot)\!\big(\psi_{\zeta}([\vect{z}_u\concat\vect{z}_v])\big).
\end{equation}

The default pre-training loss is binary cross-entropy,
\begin{equation}
    \set{L}_{\mathrm{SP}}=-\sum_{(u,v)\in\Omega_{\mathrm{tr}}}\left[y_{uv}\log \hat{y}_{uv}+(1-y_{uv})\log(1-\hat{y}_{uv})\right].
\label{eq:signloss}
\end{equation}

\subsection{Cluster prompts for downstream adaptation}
Due to the limited node classification labels in the signed graph learning literature, we focus on the task of node clustering, where abundant synthetic datasets are available~\cite{he2024pytorch}. To adapt the GPPT~\cite{sun2022gppt} idea to this setting, we replace class prompts with cluster prompts. Let $M$ be the number of METIS~\citep{karypis1998fast} clusters (note these clusters typically differ from our downstream $K$ clusters, but are constructed since varying task tokens in different densely-connected clusters may provide enhanced task embeddings). We introduce a learnable prompt bank
\begin{equation}
    \vect{P}=\{\vect{p}_1,\ldots,\vect{p}_M\},
    \qquad
    \vect{p}_m\in\R^{d_P}.
\end{equation}
For node $u$ (belonging to a certain METIS cluster $m$) and a downstream cluster $k$, we construct a prompted token
\begin{equation}
    \boldsymbol{\xi}_{u,k}=[\vect{p}_k^m\concat\vect{z}_u],
\end{equation}
and score cluster compatibility with a small task head $g_{\phi}$,
\begin{equation}
    s(u,k)=g_{\phi}(\boldsymbol{\xi}_{u,k}),
    \qquad
    \hat{c}_u = \arg\max_{k\in\{1,\ldots,K\}} s(u,k).
\end{equation}
If a training set $\set{S}\subseteq\set{V}$ of labeled nodes is available, the adaptation objective is a negative log-likelihood (NLL) loss function as
\begin{equation}
    \set{L}_{\mathrm{cluster}} = \sum_{u\in\set{S}} \mathrm{NLL}\big(\text{LogSoftmax}(s(u,\cdot)), c_u\big),
\end{equation}
where $c_u\in\{1,\ldots,K\}$ is the seed-node cluster label. In practice, the pre-trained encoder can be frozen and only the prompt bank and task head are tuned, or a lightweight adapter can be updated jointly. 

\section{Experiments}
\label{sec:experiments}
We conduct experiments on both synthetic and real-world datasets to evaluate our proposed TopoSIGN framework against baselines and its variants. Additional results are provided in Appendix~\ref{app_sec:additional_res}, while implementation details are provided in Appendix~\ref{app_sec:implementation}.
\subsection{Datasets and evaluation metrics}
To evaluate the effectiveness of the proposed TopoSIGN framework, we conduct a comprehensive comparison on both synthetic benchmarks (SDSBM from \cite{he2022msgnn}) and real-world datasets (Rainfall and SP1500 from \cite{he2022sssnet}). Rainfall and SP1500 represent the only publicly available real-world signed graphs that provide both ground-truth node labels and sufficient scale required for self-supervised pre-training and node clustering evaluation to the best of our knowledge. Dataset statistics are provided in Appendix Table~\ref{tab:data-statistics}.
We adopt signed link prediction for self-supervised pre-training, as it serves as a foundational task to capture intrinsic signed topological patterns without label reliance; these learned structural representations are subsequently transferred via prompts to guide node clustering, a downstream task directly aligned with node ground-truth annotations. Node clustering performance is measured using the Adjusted Rand Index (ARI)~\cite{hubert1985comparing} for node clustering tasks. ARI measures the similarity between predicted node clusters and ground-truth labels, where a score of 1 indicates a perfect match, and 0 indicates random clustering, making it a robust measure for evaluating how well a model captures structural communities.

Specifically, the synthetic SDSBM \cite{he2022msgnn} graphs are generated to simulate diverse community structures and noise levels. We employ the $F_1$ meta-graph configuration to define community interactions, where a size ratio $\rho$ characterizes the imbalance between the largest and smallest blocks, and $p$ determines the overall edge density. To account for structural uncertainty, we incorporate directional noise $\gamma$ and sign flip probability $\eta$, which perturb the topology and edge signs, respectively. In our experiments, we set the SDSBM under the following specific configuration: 
\begin{itemize}[noitemsep, leftmargin=*]
    \item \textbf{Setting 1 (SDSBM-1)}:$F_1(\gamma=0.25)$, $n = 1000$, $p = 0.1$, $\rho = 1.5$, and $\eta = 0.25$.
    \item \textbf{Setting 2 (SDSBM-2)}:$F_2(\gamma=0.25)$, $n = 1000$, $p = 0.1$, $\rho = 1.5$, and $\eta = 0.25$.
    \item \textbf{Setting 3 (SDSBM-3)}:$F_1(\gamma=0.1)$, $n = 1000$, $p = 0.1$, $\rho = 1.5$, and $\eta = 0.25$.
    \item \textbf{Setting 4 (SDSBM-4)}:$F_1(\gamma=0)$, $n = 1000$, $p = 0.1$, $\rho = 1.5$, and $\eta = 0$. Note that setting 4 is a noiseless setting. 
\end{itemize}

\subsection{Baselines and experiment setups}
Our main experiments are designed to answer two research questions: (1) How effective is TopoSIGN in transferring link prediction knowledge to node clustering, compared with other pre-training and fine-tuning methods? (2) How effective is TopoSIGN in learning node representations compared with other signed GNNs if we directly apply a semi-supervised node clustering task without any pre-training?

To answer these research questions, we compare TopoSIGN against two families of representative baselines. The first family consists of graph pre-training, prompting, and foundation-model methods: GPPT~\cite{sun2022gppt}, All-in-one~\cite{sun2023all}, GraphPrompt~\cite{liu2023graphprompt}, GPF~\cite{fang2023universal}, SAMGPT~\cite{yu2025samgpt}, and TopoDIG~\cite{liang2026topologyguided}. The second family consists of signed graph clustering models: SSSNET~\cite{he2022sssnet}, SigMaNet~\cite{fiorini2023sigmanet}, MSGNN~\cite{he2022msgnn}, and DSGC~\cite{zhao2025robust}. For a fair comparison on different architectures, we employ their original hyperparameter settings but use the same loss function for all methods during pre-training and the same NLL loss for prompt learning or semi-supervised node clustering. Hyperparameter settings are provided in Appendix~\ref{app_subsec:hyperparam}.

\subsection{Results}
Table~\ref{tab:main_compare_gfms} compares our proposed TopoSIGN (TopoMSGNN with SGE backbone MSGNN~\cite{he2022msgnn} and TopoSSSNET with SGE backbone SSSNET~\cite{he2022sssnet}) against existing graph pre-training and prompt learning methods. 
TopoSIGN achieves the strongest or near-strongest performance on the noisy SDSBM settings. On SDSBM-1, SDSBM-2, and SDSBM-3, TopoSSSNET obtains the best ARI among all pre-training and prompt learning methods, while TopoMSGNN is consistently competitive and improves over TopoDIG. This comparison is particularly informative because TopoDIG also contains a topological branch, but it was designed for directed unsigned graphs. The improvement of TopoSIGN over TopoDIG indicates that simply transferring a directed-graph topological pre-training framework to signed graphs is not enough; the filtration and structural encoder must be adapted to preserve sign information. The results on SDSBM-4 reveal a different pattern. SDSBM-4 is noiseless, and several unsigned prompt learning baselines perform well, especially SAMGPT, GPF, Gprompt, and All-in-one. This suggests that when the clustering signal is clean and strong, generic graph pre-training can already recover useful community structure, even without explicit signed topology. In contrast, TopoSIGN is more advantageous in the noisy signed settings, where edge signs and higher-order signed connectivity patterns are harder to exploit through standard message passing alone. This supports the main motivation of TopoSIGN: the topological branch is most useful when local signed neighborhoods are perturbed and when transferable representations must capture more stable structural regularities.

The real-world datasets are more challenging. This may be due to the dense and highly structured nature of these real-world graphs, the mismatch between link-sign pre-training and the final clustering objective, and the fact that the current signed degree-vector filtration captures only one type of signed topology. Nevertheless, TopoSIGN remains competitive on Rainfall and shows that the proposed framework can be applied beyond synthetic signed block models. The weaker SP1500 results suggest that future work should consider richer signed-directed filtrations, edge-feature-aware topological summaries, and more adaptive prompt construction for dense financial networks.

\vspace{-3mm}
\begin{table}[htbp]
\centering
\caption{Pre-training and prompt learning performance comparison in terms of downstream node clustering ARI. We report the mean ARI among five runs $\pm$ standard deviation. The best method is marked in \textbf{bold} while the second best is marked with \underline{underline}.}
\label{tab:main_compare_gfms}
\resizebox{\textwidth}{!}{%
\begin{tabular}{lcccccc}
\toprule
\textbf{Method} & \textbf{SDSBM-1} & \textbf{SDSBM-2} & \textbf{SDSBM-3} & \textbf{SDSBM-4} & \textbf{Rainfall} & \textbf{SP1500} \\
\midrule
GPPT & $0.000 \pm 0.000$ & $0.000 \pm 0.000$ & $0.000 \pm 0.000$ & $0.000 \pm 0.000$ & $0.000 \pm 0.000$ & $0.000 \pm 0.000$ \\
Gprompt & $0.026 \pm 0.027$ & $0.018 \pm 0.030$ & $0.169 \pm 0.141$ & $0.401 \pm 0.201$ & $\mathbf{0.160 \pm 0.047}$ & $0.066 \pm 0.033$ \\
GPF & $0.016 \pm 0.015$ & $0.026 \pm 0.023$ & $0.132 \pm 0.116$ & \underline{$0.403 \pm 0.202$} & $0.092 \pm 0.075$ & $\mathbf{0.081 \pm 0.011}$ \\
All-in-one & $0.015 \pm 0.016$ & $0.036 \pm 0.034$ & $0.148 \pm 0.099$ & $0.403 \pm 0.199$ & \underline{$0.124 \pm 0.024$} & $0.046 \pm 0.038$ \\
SAMGPT & $0.008 \pm 0.009$ & $0.037 \pm 0.044$ & $0.122 \pm 0.096$ & $\mathbf{0.477 \pm 0.045}$ & $0.090 \pm 0.074$ & \underline{$0.078 \pm 0.006$} \\
TopoDIG & $0.030 \pm 0.022$ & $0.094 \pm 0.033$ & $0.307 \pm 0.057$ & $0.162 \pm 0.017$ & $0.000 \pm 0.000$ & $0.010 \pm 0.017$ \\
\midrule
\textbf{TopoMSGNN (Ours)} & \underline{$0.032 \pm 0.030$} & \underline{$0.107 \pm 0.009$} & \underline{$0.372 \pm 0.042$} & $0.164 \pm 0.025$ & $0.075 \pm 0.061$ & $0.000 \pm 0.000$ \\
\textbf{TopoSSSNET (Ours)} & $\mathbf{0.076 \pm 0.031}$ & $\mathbf{0.125 \pm 0.011}$ & $\mathbf{0.379 \pm 0.036}$ & $0.157 \pm 0.023$ & $0.062 \pm 0.057$ & $0.000 \pm 0.000$ \\
\bottomrule
\end{tabular}%
}
\end{table}


Table \ref{tab:main_results_other_signed_GNNs} evaluates the topological branch in semi-supervised clustering. Adding topological features consistently boosts several signed encoders. The gain is most pronounced for SSSNET across all datasets—particularly SDSBM-1–3, Rainfall, and SP1500—suggesting that persistence-image features complement its signed propagation by capturing multiscale local connectivity and cycle structures. Extensive evaluations across various backbone-prompt combinations (Appendix Tables \ref{tab:full_comparison_with topo}--\ref{tab:full_ablation_topo_difference}) and low-data regimes (1, 3, and 5 shots; Appendix Tables \ref{tab:1_shot_comparison_with_topo}--\ref{tab:ablation_5shot_topo_difference}) further confirm TopoSIGN's efficacy, especially under extreme label scarcity and on complex real-world graphs. Notably, TopoSIGN slightly degrades DSGC, likely because DSGC's topology-optimizing pre-processing makes additional topological features redundant or noisy.

\vspace{-3mm}
\begin{table}[htbp]
\centering
\caption{Semi-supervised node clustering performance (ARI) on real-world and synthetic signed datasets. We compare our approach against state-of-the-art signed GNNs.} 
\label{tab:main_results_other_signed_GNNs}
\resizebox{\textwidth}{!}{%
\begin{tabular}{lcccccc}
\toprule
\textbf{Method} & \textbf{SDSBM-1} & \textbf{SDSBM-2} & \textbf{SDSBM-3} & \textbf{SDSBM-4} & \textbf{Rainfall} & \textbf{SP1500} \\
\midrule
DSGC & $0.134 \pm 0.231$ & $0.184 \pm 0.143$ & $0.197 \pm 0.147$ & $0.480 \pm 0.245$ & $0.102 \pm 0.056$ & $0.093 \pm 0.012$ \\
SigMaNet & $0.333 \pm 0.080$ & $0.230 \pm 0.121$ & $0.226 \pm 0.141$ & $0.705 \pm 0.139$ & $0.236 \pm 0.130$ & $0.075 \pm 0.040$ \\
MSGNN & $0.703 \pm 0.047$ & $0.714 \pm 0.012$ & $0.790 \pm 0.014$ & $0.988 \pm 0.008$ & $0.283 \pm 0.183$ & $0.116 \pm 0.050$ \\
SSSNET & $0.180 \pm 0.244$ & $0.052 \pm 0.065$ & $0.218 \pm 0.178$ & $0.431 \pm 0.387$ & $0.118 \pm 0.236$ & $0.009 \pm 0.014$ \\
\midrule
DSGC+Topo & \underline{$0.762 \pm 0.133$} & \underline{$0.777 \pm 0.160$} & \underline{$0.911 \pm 0.083$} & $\mathbf{0.989 \pm 0.009}$ & \underline{$0.472 \pm 0.044$} & $0.156 \pm 0.024$ \\
SigMaNet+Topo & $0.169 \pm 0.015$ & $0.248 \pm 0.062$ & $0.284 \pm 0.034$ & $0.458 \pm 0.234$ & $0.174 \pm 0.042$ & $0.072 \pm 0.006$ \\
MSGNN+Topo & $0.684 \pm 0.022$ & $0.559 \pm 0.105$ & $0.674 \pm 0.054$ & \underline{$0.974 \pm 0.022$} & $0.435 \pm 0.084$ & \underline{$0.170 \pm 0.013$} \\
SSSNET+Topo & $\mathbf{0.923 \pm 0.030}$ & $\mathbf{0.923 \pm 0.029}$ & $\mathbf{0.937 \pm 0.047}$ & $0.967 \pm 0.028$ & $\mathbf{0.481 \pm 0.034}$ & $\mathbf{0.298 \pm 0.041}$ \\
\bottomrule
\end{tabular}%
}
\end{table}

\subsection{Ablation study}
To demonstrate the efficacy of each component of our framework, we perform an ablation study to isolate the contribution of each component. From Table~\ref{tab:ablation_filtration} in Appendix~\ref{app_sec:additional_res}, we conclude that without the topological branch or without SGE, the performance typically drops (especially for MSGNN). Using a signed filtration is better than a positive-sign subgraph only topological baseline. Further results and discussions on different pre-training tasks, other prompting functions, as well as additional SGE compatibility, are provided in Appendix~\ref{app_sec:additional_res}.

\subsection{Computational Cost}
We further analyze the computational cost of TopoSIGN by separating the one-time topological preprocessing from model training and evaluation. Persistence images are extracted offline from node-level ego-networks and can be cached and reused across subsequent training and evaluation runs. Although this preprocessing introduces an additional upfront cost, the node-wise computation is naturally parallelizable, and incorporating the precomputed topological features incurs only moderate overhead during model training compared with the corresponding signed GNN backbones. The preprocessing cost becomes more pronounced for dense graphs such as SP1500, suggesting that more scalable topological feature extraction is an important direction for future work. Detailed preprocessing and training runtimes are reported in Appendix \ref{app_subsec:runtime}.

\subsection{Parameter Sensitivity}
We also evaluate the sensitivity of TopoSIGN to two key design parameters of the topological branch: the ego-network radius and the persistence-image resolution. Overall, the prompt-learning results remain relatively stable when increasing the neighborhood radius from 2 to 3 hops and across different persistence-image resolutions, while the semi-supervised setting exhibits greater sensitivity to the image resolution. A larger neighborhood can improve performance for several signed GNN backbones but requires more expensive topological preprocessing, whereas overly fine or coarse persistence-image discretizations may respectively introduce sparse/noisy representations or lose fine-grained topological information. These observations support our default choices of a 2-hop ego-network and a pixel size of 1.0 as practical settings that balance performance, robustness, and computational cost. Full sensitivity results are provided in Appendix \ref{app_subsec:sensitivity_khop} and \ref{app_subsec:sensitivity_pixel_size}.

\section{Conclusion}
\label{sec:conclusion}
\looseness=-1 We introduced TopoSIGN, a pioneering topology-guided pre-training and prompt-learning framework for signed graphs. TopoSIGN combines signed GNN structural embeddings with persistent-homology features from signed Dowker filtrations, pre-trains on link sign prediction, and adapts to node clustering with cluster prompts. TopoSIGN is compatible with various signed GNN backbones and prompt functions. One key methodological component is a signed degree-vector filtration that retains positive and negative degree information while remaining compatible with standard one-parameter persistence. Future work includes richer signed-directed filtrations based on separate in/out positive and negative degrees, dynamic signed graph pre-training, edge-feature-aware signed topology, and scalable landmark selection for large signed networks. We do not envision severe negative societal impacts.

\section{Acknowledgments}

Yixuan He is supported by a Jetstream2 NAIRR AI Fellowship. This work used Jetstream2 at Indiana University~\cite{hancock2021jetstream2} through allocation CIS251396 from the Advanced Cyberinfrastructure Coordination Ecosystem: Services \& Support (ACCESS) program~\cite{boerner2023access}, which is supported by National Science Foundation grants \#2138259, \#2138286, \#2138307, \#2137603, and \#2138296.

\bibliographystyle{unsrtnat}
\bibliography{reference}

\appendix
\section{Additional results}
\label{app_sec:additional_res}

Here we provide additional ablation study results. Note that results on the same method may be slightly different due to different ways of setting random seeds in the scripts.

\subsection{Analysis of Topological Components and Backbones }

Table \ref{tab:ablation_filtration} ablates the two-branch TopoSIGN design under the signed link-prediction pre-training and prompt-learning clustering pipeline. The topology-only variant yields near-zero ARI, showing that persistent-homology features alone cannot replace a signed graph encoder. For the MSGNN backbone, TopoMSGNN improves over MSGNN-only on four of the five datasets and generally outperforms the positive-sign subgraph-only topological baseline, indicating that the signed degree-vector filtration provides useful complementary information and that negative edges should not be discarded before topological feature construction. For the SSSNET backbone, the effect is more mixed: TopoSSSNET improves on SDSBM-2 but underperforms SSSNET-only on several other datasets, suggesting that topological augmentation is backbone-dependent and may interact differently with signed aggregation mechanisms. Overall, the ablation confirms that the structural encoder is essential, the topology representation learning with signed filtration is generally preferable to positive-only topology, and the gain from topology depends on how well the topological features complement the chosen signed graph encoder.

\begin{table}[htbp]
\centering
\caption{Ablation study: effect of different topological filtration strategies on node clustering ARI on TopoMSGNN and TopoSSSNET. posTopo means the positive-subgraph only baseline filtration. MSGNN-, SSSNET-, or Topo- only means leaving out one branch in the framework.} 
\label{tab:ablation_filtration}
\resizebox{\textwidth}{!}{%
\begin{tabular}{lcccccc}
\toprule
\textbf{Method} & \textbf{SDSBM-1} & \textbf{SDSBM-2} & \textbf{SDSBM-3} & \textbf{SDSBM-4} & \textbf{Rainfall} \\
\midrule
MSGNN\_only & $0.014 \pm 0.011$ & \underline{$0.102 \pm 0.021$} & $0.372 \pm 0.044$ & \underline{$0.164 \pm 0.019$} & \underline{$0.068 \pm 0.058$} \\
Topo\_only & $-0.001 \pm 0.001$ & $0.003 \pm 0.007$ & $0.007 \pm 0.008$ & $-0.000 \pm 0.000$ & $0.001 \pm 0.001$ \\
MSGNN+posTopo & $0.006 \pm 0.006$ & $0.099 \pm 0.016$ & $0.358 \pm 0.049$ & $0.164 \pm 0.034$ & $0.034 \pm 0.046$  \\
TopoMSGNN & $0.037 \pm 0.026$ & $0.092 \pm 0.018$ & \underline{$0.381 \pm 0.046$} & $\mathbf{0.167 \pm 0.024}$ & $\mathbf{0.076 \pm 0.063}$  \\
SSSNET\_only & $\mathbf{0.096 \pm 0.046}$ & $0.099 \pm 0.025$ & $\mathbf{0.394 \pm 0.019}$ & $0.162 \pm 0.023$ & $0.054 \pm 0.055$  \\
SSSNET+posTopo & $0.057 \pm 0.016$ & $0.094 \pm 0.028$ & $0.365 \pm 0.028$ & $0.148 \pm 0.021$ & $0.056 \pm 0.052$\\
TopoSSSNET & \underline{$0.066 \pm 0.039$} & $\mathbf{0.118 \pm 0.020}$ & $0.374 \pm 0.017$ & $0.146 \pm 0.028$ & $0.025 \pm 0.049$\\
\bottomrule
\end{tabular}%
}
\end{table}

In addition, we are interested in how different pre-training tasks may affect the quality of TopoSIGN's embeddings useful for node clustering. Table~\ref{tab:ablation_pretrain} provides a comparison of conducting one additional link prediction task.
For signed directed graphs, we additionally consider multi-task variants in our ablation study
\begin{equation}
    \set{L}_{\mathrm{pre}}=\set{L}_{\mathrm{SP}}+\sum_{r\in\set{R}}\alpha_r\set{L}_r,
\end{equation}
where $\set{R}$ may include DP, 3C, 4C, and 5C relation-prediction objectives and $\alpha_r$ are hyperparameters (here we set them to be either 0 or 1). These auxiliary tasks test whether directional or non-edge discrimination improves downstream clustering transfer. In addition to our default task SP, in our second task, direction prediction (DP), one aims to predict whether  $(u, v)\in\mathcal{E}$ or $(v, u)\in\mathcal{E}$ under the assumption that exactly one of these two conditions holds. We also consider three-, four-, and five-class prediction problems. In the three-class problem (3C), the possibilities are $(u, v)\in\mathcal{E},$ $(v, u)\in\mathcal{E},$ or that neither $(u, v)$ nor $(v, u)$ are in $\mathcal{E}$. For the four-class problem (4C), the possibilities are $(u, v)\in\mathcal{E}^+$, $(u, v)\in\mathcal{E}^-$, $(v, u)\in\mathcal{E}^+$, and $(v, u)\in\mathcal{E}^-$. For the five-class problem (5C), we also add in the possibility that neither $(u, v)$ nor $(v, u)$ are in $\mathcal{E}$. In this ablation study, we test SP only against SP plus one of the other tasks. We can see from Table~\ref{tab:ablation_pretrain} that adding the pre-training task 3C is generally the most helpful for TopoMSGNN by also considering a class of non-existing edges to further learn about the graph structure. On the other hand, SP only seems to perform the second-best overall. On TopoSSSNET, however, different additions to SP may have different leading datasets, and having an extra task does boost the TopoSSSNET performance. For simplicity and due to the unclearness of which extra task is the most helpful, we stick to SP only for our main results.

\begin{table}[htbp]
\centering
\caption{Pre-training task ablation: effect of different pre-training objectives on downstream node clustering ARI after prompt learning for TopoMSGNN. We report the mean ARI among five runs plus/minus one standard deviation. The best method is marked in \textbf{bold} while the second best is marked with \underline{underline}.}
\label{tab:ablation_pretrain}
\resizebox{\textwidth}{!}{%
\begin{tabular}{lcccc}
\toprule
\textbf{Method} & \textbf{SDSBM-1} & \textbf{SDSBM-2} & \textbf{SDSBM-3} & \textbf{SDSBM-4} \\
\midrule
TopoMSGNN\_pretrain\_SP & $0.013 \pm 0.025$ & $0.100 \pm 0.015$ & $0.400 \pm 0.025$ & $0.164 \pm 0.018$ \\
TopoMSGNN\_pretrain\_SP+DP & $0.026 \pm 0.006$ & $0.090 \pm 0.026$ & $0.340 \pm 0.080$ & $0.029 \pm 0.028$ \\
TopoMSGNN\_pretrain\_SP+3C & $0.031 \pm 0.024$ & $0.115 \pm 0.015$ & $0.396 \pm 0.043$ & $0.117 \pm 0.035$ \\
TopoMSGNN\_pretrain\_SP+4C & $0.011 \pm 0.010$ & $0.113 \pm 0.014$ & $0.389 \pm 0.066$ & $0.074 \pm 0.043$ \\
TopoMSGNN\_pretrain\_SP+5C & $0.013 \pm 0.014$ & $0.107 \pm 0.011$ & $0.393 \pm 0.040$ & $0.072 \pm 0.017$ \\
TopoSSSNET\_pretrain\_SP & $0.066 \pm 0.039$ & $0.118 \pm 0.020$ & $0.374 \pm 0.017$ & $0.146 \pm 0.028$ \\
TopoSSSNET\_pretrain\_SP+DP & $\mathbf{0.082 \pm 0.017}$ & \underline{$0.129 \pm 0.004$} & $0.389 \pm 0.060$ & $0.064 \pm 0.026$ \\
TopoSSSNET\_pretrain\_SP+3C & \underline{$0.072 \pm 0.038$} & $0.126 \pm 0.020$ & $\mathbf{0.431 \pm 0.023}$ & $0.189 \pm 0.150$ \\
TopoSSSNET\_pretrain\_SP+4C & $0.060 \pm 0.048$ & $\mathbf{0.132 \pm 0.015}$ & $0.399 \pm 0.090$ & \underline{$0.195 \pm 0.118$} \\
TopoSSSNET\_pretrain\_SP+5C & $0.068 \pm 0.025$ & $0.125 \pm 0.016$ & \underline{$0.430 \pm 0.039$} & $\mathbf{0.205 \pm 0.101}$ \\
\bottomrule
\end{tabular}%
}
\end{table}

\subsection{Evaluation on Graph Prompting Paradigm}

To further investigate the compatibility of our signed topological features with the emerging graph prompting paradigm, we evaluate four representative downstream prompting methods: GPPT, Gprompt, GPF, and All-in-One. In this experiment, the SGE+Topo part is employed as a fixed backbone to provide structural embeddings, and we compare how different prompt designs affect the final node clustering results. Comparing Table~\ref{tab:full_comparison_with topo} and Table~\ref{tab:main_compare_gfms}, we conclude that using our SGE+Topo indeed boosts the prompt learning frameworks in the signed clustering task.

\subsection{Extended Ablation Studies and Few-Shot Experiments}

To investigate whether our topological filtration strategies universally enhance baseline performance, we conduct comprehensive ablation studies across all combinations of backbones and downstream prompting frameworks, evaluating models both with and without topological information. We first evaluate overall performance under standard settings, with results reported in Tables~\ref{tab:full_comparison_with topo}--\ref{tab:full_ablation_topo_difference}. Furthermore, to assess whether topological features remain effective under data-scarce scenarios, we conduct few-shot ablation experiments under 1-, 3-, and 5-shot settings. The detailed few-shot results are summarized in Tables~\ref{tab:1_shot_comparison_with_topo}--\ref{tab:ablation_5shot_topo_difference}.

\begin{table}[htbp]
\centering
\caption{Prompt method comparison: performance of TopoMSGNN, TopoSSSNET, TopoDSGC and TopoSigMaNet combined with different graph prompting strategies. We report the mean ARI among five runs plus/minus one standard deviation. The best method is marked in \textbf{bold} while the second best is marked with \underline{underline}.}
\label{tab:full_comparison_with topo}
\resizebox{\textwidth}{!}{%
\begin{tabular}{lcccccc}
\toprule
\textbf{Method} & \textbf{SDSBM-1} & \textbf{SDSBM-2} & \textbf{SDSBM-3} & \textbf{SDSBM-4} & \textbf{Rainfall} & \textbf{SP1500} \\
\midrule
MSGNN+Topo+GPPT & $0.032 \pm 0.029$ & $0.086 \pm 0.032$ & $0.399 \pm 0.041$ & $0.233 \pm 0.131$ & $0.079 \pm 0.065$ & $0.005 \pm 0.010$ \\
MSGNN+Topo+Gprompt & $0.670 \pm 0.038$ & $0.680 \pm 0.031$ & $0.768 \pm 0.047$ & $0.641 \pm 0.013$ & $0.129 \pm 0.033$ & $0.095 \pm 0.012$ \\
MSGNN+Topo+GPF & $0.679 \pm 0.041$ & $0.671 \pm 0.042$ & $0.767 \pm 0.028$ & $\mathbf{0.876 \pm 0.153}$ & $0.142 \pm 0.023$ & $0.103 \pm 0.016$ \\
MSGNN+Topo+All-in-one & $0.658 \pm 0.057$ & $0.660 \pm 0.032$ & $0.748 \pm 0.050$ & $0.801 \pm 0.157$ & $0.159 \pm 0.027$ & $0.098 \pm 0.006$ \\
SSSNET+Topo+GPPT & $0.087 \pm 0.029$ & $0.139 \pm 0.019$ & $0.375 \pm 0.042$ & $0.156 \pm 0.017$ & $0.055 \pm 0.055$ & $0.000 \pm 0.000$ \\
SSSNET+Topo+Gprompt & $0.665 \pm 0.057$ & \underline{$0.792 \pm 0.057$} & $0.803 \pm 0.070$ & $0.651 \pm 0.000$ & $0.145 \pm 0.057$ & $0.108 \pm 0.019$ \\
SSSNET+Topo+GPF & $0.707 \pm 0.040$ & $0.754 \pm 0.045$ & $0.795 \pm 0.048$ & $0.720 \pm 0.107$ & $0.143 \pm 0.025$ & $\mathbf{0.119 \pm 0.030}$ \\
SSSNET+Topo+All-in-one & $0.648 \pm 0.034$ & $0.774 \pm 0.055$ & $0.818 \pm 0.056$ & $0.706 \pm 0.119$ & $0.161 \pm 0.039$ & \underline{$0.117 \pm 0.025$} \\
DSGC+Topo+GPPT & $0.042 \pm 0.037$ & $0.104 \pm 0.022$ & $0.261 \pm 0.051$ & $0.154 \pm 0.017$ & $0.025 \pm 0.049$ & $0.000 \pm 0.000$ \\
DSGC+Topo+Gprompt & \underline{$0.950 \pm 0.010$} & $\mathbf{0.826 \pm 0.035}$ & \underline{$0.897 \pm 0.075$} & $0.779 \pm 0.162$ & $0.161 \pm 0.029$ & $0.091 \pm 0.008$ \\
DSGC+Topo+GPF & $0.937 \pm 0.021$ & $0.767 \pm 0.086$ & $\mathbf{0.911 \pm 0.043}$ & \underline{$0.845 \pm 0.186$} & \underline{$0.173 \pm 0.014$} & $0.093 \pm 0.004$ \\
DSGC+Topo+All-in-one & $\mathbf{0.957 \pm 0.009}$ & $0.785 \pm 0.018$ & $0.892 \pm 0.064$ & $0.790 \pm 0.172$ & $\mathbf{0.175 \pm 0.030}$ & $0.093 \pm 0.009$ \\
SigMaNet+Topo+GPPT & $-0.001 \pm 0.002$ & $0.012 \pm 0.007$ & $0.079 \pm 0.037$ & $0.204 \pm 0.135$ & $0.033 \pm 0.019$ & $0.006 \pm 0.004$ \\
SigMaNet+Topo+Gprompt & $0.056 \pm 0.026$ & $0.050 \pm 0.038$ & $0.094 \pm 0.050$ & $0.162 \pm 0.091$ & $0.104 \pm 0.044$ & $0.030 \pm 0.017$ \\
SigMaNet+Topo+GPF & $0.090 \pm 0.059$ & $0.133 \pm 0.033$ & $0.143 \pm 0.039$ & $0.214 \pm 0.114$ & $0.134 \pm 0.030$ & $0.051 \pm 0.004$ \\
SigMaNet+Topo+All-in-one & $0.077 \pm 0.061$ & $0.118 \pm 0.030$ & $0.141 \pm 0.033$ & $0.294 \pm 0.020$ & $0.123 \pm 0.040$ & $0.036 \pm 0.011$ \\
\bottomrule
\end{tabular}%
}
\end{table}

\begin{table}[htbp]
\centering
\caption{Prompt method comparison: performance of MSGNN, SSSNET, DSGC and SigMaNet combined with different graph prompting strategies. We report the mean ARI among five runs plus/minus one standard deviation. The best method is marked in \textbf{bold} while the second best is marked with \underline{underline}.}
\label{tab:full_comparison_without_topo}
\resizebox{\textwidth}{!}{%
\begin{tabular}{lcccccc}
\toprule
\textbf{Method} & \textbf{SDSBM-1} & \textbf{SDSBM-2} & \textbf{SDSBM-3} & \textbf{SDSBM-4} & \textbf{Rainfall} & \textbf{SP1500} \\
\midrule
MSGNN+GPPT & $0.012 \pm 0.013$ & $0.104 \pm 0.016$ & $0.368 \pm 0.048$ & $0.157 \pm 0.016$ & $0.073 \pm 0.060$ & $0.010 \pm 0.012$ \\
MSGNN+Gprompt & $0.640 \pm 0.099$ & $0.643 \pm 0.117$ & $0.764 \pm 0.045$ & $0.768 \pm 0.132$ & $0.151 \pm 0.023$ & $0.105 \pm 0.008$ \\
MSGNN+GPF & $0.676 \pm 0.016$ & $0.688 \pm 0.019$ & $0.756 \pm 0.022$ & $0.758 \pm 0.134$ & $0.161 \pm 0.027$ & $0.103 \pm 0.015$ \\
MSGNN+All-in-one & $0.660 \pm 0.032$ & $0.705 \pm 0.016$ & $0.787 \pm 0.022$ & $0.787 \pm 0.167$ & $0.169 \pm 0.022$ & \underline{$0.107 \pm 0.007$} \\
SSSNET+GPPT & $0.057 \pm 0.026$ & $0.120 \pm 0.017$ & $0.382 \pm 0.041$ & $0.164 \pm 0.018$ & $0.053 \pm 0.065$ & $0.000 \pm 0.000$ \\
SSSNET+Gprompt & $0.654 \pm 0.023$ & $0.779 \pm 0.065$ & $0.925 \pm 0.049$ & $0.716 \pm 0.130$ & $0.126 \pm 0.027$ & $0.096 \pm 0.015$ \\
SSSNET+GPF & $0.694 \pm 0.054$ & $0.773 \pm 0.023$ & $0.899 \pm 0.074$ & $0.749 \pm 0.130$ & $0.148 \pm 0.020$ & $0.102 \pm 0.012$ \\
SSSNET+All-in-one & $0.710 \pm 0.043$ & $0.775 \pm 0.041$ & $0.867 \pm 0.054$ & $0.805 \pm 0.143$ & $\mathbf{0.178 \pm 0.051}$ & $\mathbf{0.115 \pm 0.028}$ \\
DSGC+GPPT & $0.057 \pm 0.029$ & $0.106 \pm 0.007$ & $0.291 \pm 0.062$ & $0.210 \pm 0.155$ & $0.026 \pm 0.053$ & $0.000 \pm 0.000$ \\
DSGC+Gprompt & \underline{$0.927 \pm 0.049$} & $\mathbf{0.866 \pm 0.041}$ & $\mathbf{0.957 \pm 0.010}$ & $0.791 \pm 0.171$ & $0.164 \pm 0.046$ & $0.096 \pm 0.006$ \\
DSGC+GPF & $0.909 \pm 0.035$ & $0.828 \pm 0.038$ & $0.948 \pm 0.023$ & \underline{$0.858 \pm 0.169$} & \underline{$0.170 \pm 0.031$} & $0.084 \pm 0.013$ \\
DSGC+All-in-one & $\mathbf{0.941 \pm 0.023}$ & \underline{$0.857 \pm 0.032$} & \underline{$0.956 \pm 0.010$} & $\mathbf{0.860 \pm 0.171}$ & $0.155 \pm 0.025$ & $0.089 \pm 0.007$ \\
SigMaNet+GPPT & $-0.001 \pm 0.004$ & $0.006 \pm 0.009$ & $0.028 \pm 0.051$ & $0.082 \pm 0.165$ & $0.060 \pm 0.121$ & $0.001 \pm 0.001$ \\
SigMaNet+Gprompt & $0.114 \pm 0.061$ & $0.044 \pm 0.088$ & $0.186 \pm 0.130$ & $0.274 \pm 0.357$ & $0.107 \pm 0.148$ & $0.018 \pm 0.028$ \\
SigMaNet+GPF & $0.097 \pm 0.101$ & $0.157 \pm 0.124$ & $0.233 \pm 0.131$ & $0.664 \pm 0.118$ & $-0.000 \pm 0.000$ & $0.043 \pm 0.035$ \\
SigMaNet+All-in-one & $0.044 \pm 0.061$ & $0.093 \pm 0.115$ & $0.222 \pm 0.119$ & $0.466 \pm 0.206$ & $0.064 \pm 0.079$ & $0.033 \pm 0.039$ \\
\bottomrule
\end{tabular}%
}
\end{table}

\begin{table}[htbp]
\centering
\caption{Ablation study: performance improvement by adding topological information (Table 5 minus Table 6). Positive values indicate performance gains due to topological structures, while negative values show performance drops.}
\label{tab:full_ablation_topo_difference}
\resizebox{\textwidth}{!}{%
\begin{tabular}{lcccccc}
\toprule
\textbf{Method} & \textbf{SDSBM-1} & \textbf{SDSBM-2} & \textbf{SDSBM-3} & \textbf{SDSBM-4} & \textbf{Rainfall} & \textbf{SP1500} \\
\midrule
MSGNN+GPPT & $+0.020$ & $-0.018$ & $+0.031$ & $+0.076$ & $+0.006$ & $-0.005$ \\
MSGNN+Gprompt & $+0.030$ & $+0.037$ & $+0.004$ & $-0.127$ & $-0.022$ & $-0.010$ \\
MSGNN+GPF & $+0.003$ & $-0.017$ & $+0.011$ & $+0.118$ & $-0.019$ & $0.000$ \\
MSGNN+All-in-one & $-0.002$ & $-0.045$ & $-0.039$ & $+0.014$ & $-0.010$ & $-0.009$ \\
\midrule
SSSNET+GPPT & $+0.030$ & $+0.019$ & $-0.007$ & $-0.008$ & $+0.002$ & $0.000$ \\
SSSNET+Gprompt & $+0.011$ & $+0.013$ & $-0.122$ & $-0.065$ & $+0.019$ & $+0.012$ \\
SSSNET+GPF & $+0.013$ & $-0.019$ & $-0.104$ & $-0.029$ & $-0.005$ & $+0.017$ \\
SSSNET+All-in-one & $-0.062$ & $-0.001$ & $-0.049$ & $-0.099$ & $-0.017$ & $+0.002$ \\
\midrule
DSGC+GPPT & $-0.015$ & $-0.002$ & $-0.030$ & $-0.056$ & $-0.001$ & $0.000$ \\
DSGC+Gprompt & $+0.023$ & $-0.040$ & $-0.060$ & $-0.012$ & $-0.003$ & $-0.005$ \\
DSGC+GPF & $+0.028$ & $-0.061$ & $-0.037$ & $-0.013$ & $+0.003$ & $+0.009$ \\
DSGC+All-in-one & $+0.016$ & $-0.072$ & $-0.064$ & $-0.070$ & $+0.020$ & $+0.004$ \\
\midrule
SigMaNet+GPPT & $0.000$ & $+0.006$ & $+0.051$ & $+0.122$ & $-0.027$ & $+0.005$ \\
SigMaNet+Gprompt & $-0.058$ & $+0.006$ & $-0.092$ & $-0.112$ & $-0.003$ & $+0.012$ \\
SigMaNet+GPF & $-0.007$ & $-0.024$ & $-0.090$ & $-0.450$ & $+0.134$ & $+0.008$ \\
SigMaNet+All-in-one & $+0.033$ & $+0.025$ & $-0.081$ & $-0.172$ & $+0.059$ & $+0.003$ \\
\bottomrule
\end{tabular}%
}
\end{table}

\begin{table}[htbp]
\centering
\caption{Prompt method comparison with topological information (1-shot with topo). We report the mean ARI among five runs plus/minus one standard deviation. The best method is marked in \textbf{bold} while the second best is marked with \underline{underline}.}
\label{tab:1_shot_comparison_with_topo}
\resizebox{\textwidth}{!}{%
\begin{tabular}{lcccccc}
\toprule
\textbf{Method} & \textbf{SDSBM-1} & \textbf{SDSBM-2} & \textbf{SDSBM-3} & \textbf{SDSBM-4} & \textbf{Rainfall} & \textbf{SP1500} \\
\midrule
TopoMSGNN+GPPT & $0.024 \pm 0.036$ & $0.092 \pm 0.038$ & $0.334 \pm 0.055$ & $0.158 \pm 0.034$ & $0.137 \pm 0.010$ & $0.088 \pm 0.022$ \\
TopoMSGNN+Gprompt & $0.513 \pm 0.067$ & $0.572 \pm 0.078$ & $0.593 \pm 0.159$ & $0.858 \pm 0.168$ & $0.170 \pm 0.045$ & $0.107 \pm 0.011$ \\
TopoMSGNN+GPF & $0.504 \pm 0.098$ & $0.545 \pm 0.056$ & $0.620 \pm 0.125$ & $\mathbf{0.967 \pm 0.024}$ & $0.194 \pm 0.016$ & $0.113 \pm 0.011$ \\
TopoMSGNN+All-in-one & $0.563 \pm 0.087$ & $0.541 \pm 0.052$ & $0.605 \pm 0.158$ & $0.835 \pm 0.153$ & $0.171 \pm 0.014$ & $0.113 \pm 0.012$ \\
TopoSSSNET+GPPT & $0.041 \pm 0.012$ & $0.101 \pm 0.012$ & $0.353 \pm 0.038$ & $0.160 \pm 0.025$ & $0.124 \pm 0.022$ & $0.069 \pm 0.017$ \\
TopoSSSNET+Gprompt & $0.667 \pm 0.024$ & $\mathbf{0.755 \pm 0.042}$ & $0.799 \pm 0.043$ & $0.903 \pm 0.129$ & \underline{$0.211 \pm 0.031$} & $\mathbf{0.118 \pm 0.021}$ \\
TopoSSSNET+GPF & $0.681 \pm 0.034$ & $0.738 \pm 0.016$ & $0.820 \pm 0.063$ & \underline{$0.907 \pm 0.130$} & $\mathbf{0.216 \pm 0.018}$ & $0.106 \pm 0.011$ \\
TopoSSSNET+All-in-one & $0.670 \pm 0.035$ & \underline{$0.742 \pm 0.049$} & $0.817 \pm 0.045$ & $0.892 \pm 0.124$ & $0.168 \pm 0.036$ & \underline{$0.115 \pm 0.010$} \\
TopoDSGC+GPPT & $0.007 \pm 0.011$ & $0.085 \pm 0.022$ & $0.273 \pm 0.054$ & $0.156 \pm 0.022$ & $0.124 \pm 0.028$ & $0.082 \pm 0.003$ \\
TopoDSGC+Gprompt & \underline{$0.935 \pm 0.015$} & $0.668 \pm 0.121$ & \underline{$0.850 \pm 0.073$} & $0.835 \pm 0.147$ & $0.180 \pm 0.024$ & $0.100 \pm 0.008$ \\
TopoDSGC+GPF & $0.933 \pm 0.015$ & $0.660 \pm 0.075$ & $0.822 \pm 0.102$ & $0.860 \pm 0.156$ & $0.189 \pm 0.016$ & $0.096 \pm 0.006$ \\
TopoDSGC+All-in-one & $\mathbf{0.944 \pm 0.017}$ & $0.701 \pm 0.153$ & $\mathbf{0.850 \pm 0.082}$ & $0.853 \pm 0.127$ & $0.168 \pm 0.013$ & $0.090 \pm 0.008$ \\
TopoSigMaNet+GPPT & $-0.001 \pm 0.003$ & $0.001 \pm 0.002$ & $0.084 \pm 0.054$ & $0.215 \pm 0.106$ & $0.040 \pm 0.031$ & $0.009 \pm 0.008$ \\
TopoSigMaNet+Gprompt & $0.076 \pm 0.013$ & $0.129 \pm 0.024$ & $0.128 \pm 0.031$ & $0.235 \pm 0.125$ & $0.133 \pm 0.022$ & $0.030 \pm 0.017$ \\
TopoSigMaNet+GPF & $0.101 \pm 0.038$ & $0.136 \pm 0.030$ & $0.152 \pm 0.025$ & $0.329 \pm 0.023$ & $0.137 \pm 0.020$ & $0.034 \pm 0.019$ \\
TopoSigMaNet+All-in-one & $0.112 \pm 0.032$ & $0.112 \pm 0.054$ & $0.128 \pm 0.032$ & $0.300 \pm 0.080$ & $0.129 \pm 0.021$ & $0.035 \pm 0.019$ \\
\bottomrule
\end{tabular}%
}
\end{table}

\begin{table}[htbp]
\centering
\caption{Prompt method comparison without topological information (1-shot without topo).}
\label{tab:1_shot_comparison_without_topo}
\resizebox{\textwidth}{!}{%
\begin{tabular}{lcccccc}
\toprule
\textbf{Method} & \textbf{SDSBM-1} & \textbf{SDSBM-2} & \textbf{SDSBM-3} & \textbf{SDSBM-4} & \textbf{Rainfall} & \textbf{SP1500} \\
\midrule
MSGNN+GPPT & $0.016 \pm 0.009$ & $0.081 \pm 0.037$ & $0.321 \pm 0.053$ & $0.166 \pm 0.029$ & $0.112 \pm 0.057$ & $0.084 \pm 0.005$ \\
MSGNN+Gprompt & $0.517 \pm 0.084$ & $0.547 \pm 0.071$ & $0.570 \pm 0.134$ & \underline{$0.952 \pm 0.051$} & $0.176 \pm 0.017$ & \underline{$0.113 \pm 0.009$} \\
MSGNN+GPF & $0.533 \pm 0.085$ & $0.534 \pm 0.060$ & $0.641 \pm 0.081$ & $0.948 \pm 0.045$ & $0.172 \pm 0.023$ & $0.113 \pm 0.016$ \\
MSGNN+All-in-one & $0.498 \pm 0.026$ & $0.564 \pm 0.057$ & $0.602 \pm 0.068$ & $\mathbf{0.980 \pm 0.014}$ & $0.178 \pm 0.014$ & $0.108 \pm 0.009$ \\
\midrule
SSSNET+GPPT & $0.035 \pm 0.018$ & $0.089 \pm 0.034$ & $0.361 \pm 0.049$ & $0.172 \pm 0.019$ & $0.134 \pm 0.015$ & $0.068 \pm 0.008$ \\
SSSNET+Gprompt & $0.648 \pm 0.036$ & $0.729 \pm 0.030$ & $0.791 \pm 0.065$ & $0.899 \pm 0.129$ & $\mathbf{0.226 \pm 0.040}$ & $0.112 \pm 0.008$ \\
SSSNET+GPF & $0.721 \pm 0.056$ & \underline{$0.755 \pm 0.064$} & $0.814 \pm 0.080$ & $0.909 \pm 0.131$ & $0.194 \pm 0.028$ & $0.104 \pm 0.009$ \\
SSSNET+All-in-one & $0.648 \pm 0.027$ & $0.711 \pm 0.062$ & $0.797 \pm 0.038$ & $0.937 \pm 0.099$ & $0.180 \pm 0.012$ & $\mathbf{0.116 \pm 0.016}$ \\
\midrule
DSGC+GPPT & $0.024 \pm 0.014$ & $0.109 \pm 0.015$ & $0.329 \pm 0.047$ & $0.202 \pm 0.159$ & $0.127 \pm 0.018$ & $0.062 \pm 0.029$ \\
DSGC+Gprompt & \underline{$0.916 \pm 0.032$} & $0.748 \pm 0.136$ & $\mathbf{0.930 \pm 0.024}$ & $0.861 \pm 0.170$ & \underline{$0.204 \pm 0.015$} & $0.100 \pm 0.006$ \\
DSGC+GPF & $\mathbf{0.920 \pm 0.031}$ & $0.724 \pm 0.078$ & \underline{$0.912 \pm 0.035$} & $0.861 \pm 0.170$ & $0.195 \pm 0.023$ & $0.095 \pm 0.003$ \\
DSGC+All-in-one & $0.881 \pm 0.031$ & $\mathbf{0.781 \pm 0.045}$ & $0.883 \pm 0.070$ & $0.860 \pm 0.163$ & $0.183 \pm 0.027$ & $0.094 \pm 0.008$ \\
\midrule
SigMaNet+GPPT & $-0.001 \pm 0.002$ & $0.009 \pm 0.013$ & $0.005 \pm 0.010$ & $0.000 \pm 0.000$ & $0.021 \pm 0.043$ & $0.015 \pm 0.024$ \\
SigMaNet+Gprompt & $0.128 \pm 0.092$ & $0.044 \pm 0.081$ & $0.188 \pm 0.155$ & $0.557 \pm 0.101$ & $0.033 \pm 0.067$ & $0.029 \pm 0.035$ \\
SigMaNet+GPF & $0.095 \pm 0.093$ & $0.108 \pm 0.132$ & $0.081 \pm 0.118$ & $0.636 \pm 0.115$ & $0.000 \pm 0.000$ & $0.040 \pm 0.040$ \\
SigMaNet+All-in-one & $0.160 \pm 0.094$ & $0.049 \pm 0.090$ & $0.201 \pm 0.166$ & $0.628 \pm 0.186$ & $0.096 \pm 0.128$ & $0.030 \pm 0.037$ \\
\bottomrule
\end{tabular}%
}
\end{table}

\begin{table}[htbp]
\centering
\caption{Ablation study on 1-shot learning: performance improvement by adding topological information (1-shot with Topo minus 1-shot without Topo). Positive values indicate gains from topology, while negative values indicate performance drops.}
\label{tab:ablation_1shot_topo_difference}
\resizebox{\textwidth}{!}{%
\begin{tabular}{lcccccc}
\toprule
\textbf{Method} & \textbf{SDSBM-1} & \textbf{SDSBM-2} & \textbf{SDSBM-3} & \textbf{SDSBM-4} & \textbf{Rainfall} & \textbf{SP1500} \\
\midrule
MSGNN+GPPT & $+0.008$ & $+0.011$ & $+0.013$ & $-0.008$ & $+0.025$ & $+0.004$ \\
MSGNN+Gprompt & $-0.004$ & $+0.025$ & $+0.023$ & $-0.094$ & $-0.006$ & $-0.006$ \\
MSGNN+GPF & $-0.029$ & $+0.011$ & $-0.021$ & $+0.019$ & $+0.022$ & $0.000$ \\
MSGNN+All-in-one & $+0.065$ & $-0.023$ & $+0.003$ & $-0.145$ & $-0.007$ & $+0.005$ \\
\midrule
SSSNET+GPPT & $+0.006$ & $+0.012$ & $-0.008$ & $-0.012$ & $-0.010$ & $+0.001$ \\
SSSNET+Gprompt & $+0.019$ & $+0.026$ & $+0.008$ & $+0.004$ & $-0.015$ & $+0.006$ \\
SSSNET+GPF & $-0.040$ & $-0.017$ & $+0.006$ & $-0.002$ & $+0.022$ & $+0.002$ \\
SSSNET+All-in-one & $+0.022$ & $+0.031$ & $+0.020$ & $-0.045$ & $-0.012$ & $-0.001$ \\
\midrule
DSGC+GPPT & $-0.015$ & $-0.024$ & $-0.056$ & $-0.046$ & $-0.003$ & $+0.020$ \\
DSGC+Gprompt & $+0.019$ & $-0.080$ & $-0.080$ & $-0.026$ & $-0.024$ & $0.000$ \\
DSGC+GPF & $+0.013$ & $-0.064$ & $-0.090$ & $-0.001$ & $-0.006$ & $+0.001$ \\
DSGC+All-in-one & $+0.063$ & $-0.080$ & $-0.033$ & $-0.007$ & $-0.015$ & $-0.004$ \\
\midrule
SigMaNet+GPPT & $0.000$ & $-0.008$ & $+0.079$ & $+0.215$ & $+0.019$ & $-0.006$ \\
SigMaNet+Gprompt & $-0.052$ & $+0.085$ & $-0.060$ & $-0.322$ & $+0.100$ & $+0.001$ \\
SigMaNet+GPF & $+0.006$ & $+0.028$ & $+0.071$ & $-0.307$ & $+0.137$ & $-0.006$ \\
SigMaNet+All-in-one & $-0.048$ & $+0.063$ & $-0.073$ & $-0.328$ & $+0.033$ & $+0.005$ \\
\bottomrule
\end{tabular}%
}
\end{table}

\begin{table}[htbp]
\centering
\caption{Prompt method comparison with topological information (3-shot with topo). We report the mean ARI among five runs plus/minus one standard deviation. The best method is marked in \textbf{bold} while the second best is marked with \underline{underline}.}
\label{tab:3_shot_comparison_with_topo}
\resizebox{\textwidth}{!}{%
\begin{tabular}{lcccccc}
\toprule
\textbf{Method} & \textbf{SDSBM-1} & \textbf{SDSBM-2} & \textbf{SDSBM-3} & \textbf{SDSBM-4} & \textbf{Rainfall} & \textbf{SP1500} \\
\midrule
TopoMSGNN+GPPT & $0.030 \pm 0.023$ & $0.081 \pm 0.017$ & $0.384 \pm 0.034$ & $0.165 \pm 0.016$ & $0.130 \pm 0.069$ & $0.099 \pm 0.007$ \\
TopoMSGNN+Gprompt & $0.543 \pm 0.059$ & $0.548 \pm 0.063$ & $0.679 \pm 0.082$ & $0.848 \pm 0.161$ & $0.182 \pm 0.007$ & $0.113 \pm 0.006$ \\
TopoMSGNN+GPF & $0.539 \pm 0.092$ & $0.528 \pm 0.027$ & $0.726 \pm 0.039$ & $\mathbf{0.954 \pm 0.042}$ & $0.178 \pm 0.036$ & $0.109 \pm 0.014$ \\
TopoMSGNN+All-in-one & $0.562 \pm 0.061$ & $0.532 \pm 0.021$ & $0.663 \pm 0.104$ & $0.842 \pm 0.157$ & $0.186 \pm 0.006$ & $0.109 \pm 0.012$ \\
TopoSSSNET+GPPT & $0.065 \pm 0.050$ & $0.118 \pm 0.024$ & $0.385 \pm 0.045$ & $0.191 \pm 0.043$ & $0.138 \pm 0.014$ & $0.067 \pm 0.011$ \\
TopoSSSNET+Gprompt & $0.671 \pm 0.027$ & \underline{$0.798 \pm 0.042$} & $0.827 \pm 0.064$ & $0.941 \pm 0.029$ & $0.194 \pm 0.019$ & $0.113 \pm 0.006$ \\
TopoSSSNET+GPF & $0.679 \pm 0.035$ & $\mathbf{0.799 \pm 0.030}$ & $0.817 \pm 0.048$ & $0.950 \pm 0.029$ & \underline{$0.219 \pm 0.057$} & \underline{$0.117 \pm 0.010$} \\
TopoSSSNET+All-in-one & $0.679 \pm 0.031$ & $0.770 \pm 0.036$ & $0.854 \pm 0.035$ & \underline{$0.950 \pm 0.025$} & $\mathbf{0.230 \pm 0.037}$ & $\mathbf{0.126 \pm 0.032}$ \\
TopoDSGC+GPPT & $0.020 \pm 0.021$ & $0.106 \pm 0.024$ & $0.271 \pm 0.092$ & $0.153 \pm 0.026$ & $0.124 \pm 0.028$ & $0.083 \pm 0.002$ \\
TopoDSGC+Gprompt & $0.926 \pm 0.017$ & $0.762 \pm 0.030$ & \underline{$0.870 \pm 0.069$} & $0.899 \pm 0.128$ & $0.183 \pm 0.025$ & $0.095 \pm 0.006$ \\
TopoDSGC+GPF & $\mathbf{0.934 \pm 0.024}$ & $0.723 \pm 0.079$ & $0.868 \pm 0.055$ & $0.857 \pm 0.155$ & $0.181 \pm 0.012$ & $0.095 \pm 0.012$ \\
TopoDSGC+All-in-one & \underline{$0.929 \pm 0.027$} & $0.724 \pm 0.042$ & $\mathbf{0.874 \pm 0.082}$ & $0.886 \pm 0.125$ & $0.179 \pm 0.021$ & $0.089 \pm 0.009$ \\
TopoSigMaNet+GPPT & $-0.004 \pm 0.003$ & $0.007 \pm 0.006$ & $0.116 \pm 0.049$ & $0.182 \pm 0.078$ & $0.028 \pm 0.022$ & $0.003 \pm 0.006$ \\
TopoSigMaNet+Gprompt & $0.086 \pm 0.039$ & $0.140 \pm 0.035$ & $0.149 \pm 0.017$ & $0.243 \pm 0.140$ & $0.126 \pm 0.035$ & $0.055 \pm 0.004$ \\
TopoSigMaNet+GPF & $0.135 \pm 0.033$ & $0.127 \pm 0.031$ & $0.162 \pm 0.029$ & $0.346 \pm 0.027$ & $0.147 \pm 0.017$ & $0.051 \pm 0.004$ \\
TopoSigMaNet+All-in-one & $0.117 \pm 0.035$ & $0.139 \pm 0.028$ & $0.148 \pm 0.020$ & $0.321 \pm 0.044$ & $0.116 \pm 0.032$ & $0.052 \pm 0.004$ \\
\bottomrule
\end{tabular}%
}
\end{table}

\begin{table}[htbp]
\centering
\caption{Prompt method comparison without topological information (3-shot without topo).}
\label{tab:ablation_3shot_no_topo_aligned}
\resizebox{\textwidth}{!}{%
\begin{tabular}{lcccccc}
\toprule
\textbf{Method} & \textbf{SDSBM-1} & \textbf{SDSBM-2} & \textbf{SDSBM-3} & \textbf{SDSBM-4} & \textbf{Rainfall} & \textbf{SP1500} \\
\midrule
MSGNN+GPPT & $0.038 \pm 0.031$ & $0.091 \pm 0.032$ & $0.360 \pm 0.039$ & $0.163 \pm 0.028$ & $0.112 \pm 0.059$ & $0.087 \pm 0.007$ \\
MSGNN+Gprompt & $0.550 \pm 0.115$ & $0.560 \pm 0.051$ & $0.719 \pm 0.063$ & $0.921 \pm 0.135$ & $0.174 \pm 0.005$ & $0.113 \pm 0.006$ \\
MSGNN+GPF & $0.578 \pm 0.045$ & $0.513 \pm 0.085$ & $0.767 \pm 0.033$ & $0.946 \pm 0.059$ & $0.173 \pm 0.019$ & $0.106 \pm 0.006$ \\
MSGNN+All-in-one & $0.602 \pm 0.072$ & $0.567 \pm 0.076$ & $0.704 \pm 0.064$ & $\mathbf{0.967 \pm 0.040}$ & $0.159 \pm 0.021$ & $0.105 \pm 0.008$ \\
\midrule
SSSNET+GPPT & $0.052 \pm 0.021$ & $0.089 \pm 0.045$ & $0.365 \pm 0.052$ & $0.169 \pm 0.019$ & $0.132 \pm 0.016$ & $0.050 \pm 0.028$ \\
SSSNET+Gprompt & $0.672 \pm 0.039$ & $0.768 \pm 0.062$ & $0.811 \pm 0.059$ & $0.901 \pm 0.133$ & $\mathbf{0.220 \pm 0.007}$ & \underline{$0.126 \pm 0.020$} \\
SSSNET+GPF & $0.724 \pm 0.056$ & $0.747 \pm 0.048$ & $0.853 \pm 0.063$ & $0.926 \pm 0.127$ & \underline{$0.208 \pm 0.020$} & $0.118 \pm 0.012$ \\
SSSNET+All-in-one & $0.647 \pm 0.060$ & $0.773 \pm 0.055$ & $0.796 \pm 0.098$ & \underline{$0.950 \pm 0.060$} & $0.201 \pm 0.035$ & $\mathbf{0.128 \pm 0.016}$ \\
\midrule
DSGC+GPPT & $0.033 \pm 0.021$ & $0.092 \pm 0.018$ & $0.344 \pm 0.026$ & $0.217 \pm 0.119$ & $0.149 \pm 0.022$ & $0.061 \pm 0.032$ \\
DSGC+Gprompt & $\mathbf{0.928 \pm 0.029}$ & $\mathbf{0.819 \pm 0.021}$ & $0.894 \pm 0.092$ & $0.861 \pm 0.170$ & $0.190 \pm 0.008$ & $0.098 \pm 0.006$ \\
DSGC+GPF & $0.914 \pm 0.031$ & $0.756 \pm 0.034$ & \underline{$0.922 \pm 0.046$} & $0.856 \pm 0.167$ & $0.189 \pm 0.028$ & $0.097 \pm 0.005$ \\
DSGC+All-in-one & \underline{$0.922 \pm 0.020$} & \underline{$0.807 \pm 0.040$} & $\mathbf{0.934 \pm 0.036}$ & $0.863 \pm 0.167$ & $0.190 \pm 0.016$ & $0.094 \pm 0.006$ \\
\midrule
SigMaNet+GPPT & $-0.001 \pm 0.002$ & $0.006 \pm 0.007$ & $0.008 \pm 0.011$ & $0.148 \pm 0.180$ & $0.003 \pm 0.007$ & $0.001 \pm 0.003$ \\
SigMaNet+Gprompt & $0.134 \pm 0.081$ & $0.048 \pm 0.086$ & $0.191 \pm 0.157$ & $0.663 \pm 0.147$ & $0.026 \pm 0.052$ & $0.023 \pm 0.029$ \\
SigMaNet+GPF & $0.094 \pm 0.091$ & $0.098 \pm 0.120$ & $0.080 \pm 0.119$ & $0.663 \pm 0.115$ & $0.000 \pm 0.000$ & $0.039 \pm 0.038$ \\
SigMaNet+All-in-one & $0.125 \pm 0.079$ & $0.046 \pm 0.087$ & $0.244 \pm 0.228$ & $0.603 \pm 0.152$ & $0.079 \pm 0.104$ & $0.031 \pm 0.038$ \\
\bottomrule
\end{tabular}%
}
\end{table}

\begin{table}[htbp]
\centering
\caption{Ablation study on 3-shot learning: performance improvement by adding topological information (3-shot with Topo minus 3-shot without Topo). Positive values represent gains from adding topological information.}
\label{tab:ablation_3shot_topo_difference}
\resizebox{\textwidth}{!}{%
\begin{tabular}{lcccccc}
\toprule
\textbf{Method} & \textbf{SDSBM-1} & \textbf{SDSBM-2} & \textbf{SDSBM-3} & \textbf{SDSBM-4} & \textbf{Rainfall} & \textbf{SP1500} \\
\midrule
MSGNN+GPPT & $-0.008$ & $-0.010$ & $+0.024$ & $+0.002$ & $+0.018$ & $+0.012$ \\
MSGNN+Gprompt & $-0.007$ & $-0.012$ & $-0.040$ & $-0.073$ & $+0.008$ & $0.000$ \\
MSGNN+GPF & $-0.039$ & $+0.015$ & $-0.041$ & $+0.008$ & $+0.005$ & $+0.003$ \\
MSGNN+All-in-one & $-0.040$ & $-0.035$ & $-0.041$ & $-0.125$ & $+0.027$ & $+0.004$ \\
\midrule
SSSNET+GPPT & $+0.013$ & $+0.029$ & $+0.020$ & $+0.022$ & $+0.006$ & $+0.017$ \\
SSSNET+Gprompt & $-0.001$ & $+0.030$ & $+0.016$ & $+0.040$ & $-0.026$ & $-0.013$ \\
SSSNET+GPF & $-0.045$ & $+0.052$ & $-0.036$ & $+0.024$ & $+0.011$ & $-0.001$ \\
SSSNET+All-in-one & $+0.032$ & $-0.003$ & $+0.058$ & $0.000$ & $+0.029$ & $-0.002$ \\
\midrule
DSGC+GPPT & $-0.013$ & $+0.014$ & $-0.073$ & $-0.064$ & $-0.025$ & $+0.022$ \\
DSGC+Gprompt & $-0.002$ & $-0.057$ & $-0.024$ & $+0.038$ & $-0.007$ & $-0.003$ \\
DSGC+GPF & $+0.020$ & $-0.033$ & $-0.054$ & $+0.001$ & $-0.008$ & $-0.002$ \\
DSGC+All-in-one & $+0.007$ & $-0.083$ & $-0.060$ & $+0.023$ & $-0.011$ & $-0.005$ \\
\midrule
SigMaNet+GPPT & $-0.003$ & $+0.001$ & $+0.108$ & $+0.034$ & $+0.025$ & $+0.002$ \\
SigMaNet+Gprompt & $-0.048$ & $+0.092$ & $-0.042$ & $-0.420$ & $+0.100$ & $+0.032$ \\
SigMaNet+GPF & $+0.041$ & $+0.029$ & $+0.082$ & $-0.317$ & $+0.147$ & $+0.012$ \\
SigMaNet+All-in-one & $-0.008$ & $+0.093$ & $-0.096$ & $-0.282$ & $+0.037$ & $+0.021$ \\
\bottomrule
\end{tabular}%
}
\end{table}

\begin{table}[htbp]
\centering
\caption{Prompt method comparison with topological information (5-shot with topo). We report the mean ARI among five runs plus/minus one standard deviation. The best method is marked in \textbf{bold} while the second best is marked with \underline{underline}.}
\label{tab:3_shot_comparison_without_topo}
\resizebox{\textwidth}{!}{%
\begin{tabular}{lcccccc}
\toprule
\textbf{Method} & \textbf{SDSBM-1} & \textbf{SDSBM-2} & \textbf{SDSBM-3} & \textbf{SDSBM-4} & \textbf{Rainfall} & \textbf{SP1500} \\
\midrule
TopoMSGNN+GPPT & $0.036 \pm 0.020$ & $0.068 \pm 0.023$ & $0.349 \pm 0.075$ & $0.160 \pm 0.024$ & $0.143 \pm 0.018$ & $0.099 \pm 0.007$ \\
TopoMSGNN+Gprompt & $0.603 \pm 0.075$ & $0.623 \pm 0.041$ & $0.691 \pm 0.076$ & $0.857 \pm 0.167$ & $0.177 \pm 0.024$ & \underline{$0.108 \pm 0.011$} \\
TopoMSGNN+GPF & $0.575 \pm 0.084$ & $0.568 \pm 0.046$ & $0.735 \pm 0.044$ & $\mathbf{0.970 \pm 0.037}$ & $0.201 \pm 0.017$ & $0.108 \pm 0.006$ \\
TopoMSGNN+All-in-one & $0.615 \pm 0.076$ & $0.590 \pm 0.071$ & $0.687 \pm 0.077$ & $0.855 \pm 0.165$ & $0.170 \pm 0.017$ & $0.102 \pm 0.007$ \\
TopoSSSNET+GPPT & $0.091 \pm 0.058$ & $0.122 \pm 0.032$ & $0.381 \pm 0.071$ & $0.151 \pm 0.025$ & $0.130 \pm 0.024$ & $0.067 \pm 0.031$ \\
TopoSSSNET+Gprompt & $0.664 \pm 0.031$ & $0.772 \pm 0.046$ & $0.855 \pm 0.044$ & $0.913 \pm 0.100$ & $0.211 \pm 0.028$ & $\mathbf{0.114 \pm 0.006}$ \\
TopoSSSNET+GPF & $0.709 \pm 0.062$ & $\mathbf{0.785 \pm 0.037}$ & $0.836 \pm 0.052$ & $0.881 \pm 0.120$ & \underline{$0.217 \pm 0.011$} & $0.096 \pm 0.015$ \\
TopoSSSNET+All-in-one & $0.680 \pm 0.042$ & $0.775 \pm 0.045$ & $0.836 \pm 0.047$ & \underline{$0.954 \pm 0.023$} & $\mathbf{0.223 \pm 0.030}$ & $\mathbf{0.114 \pm 0.009}$ \\
TopoDSGC+GPPT & $0.029 \pm 0.014$ & $0.072 \pm 0.041$ & $0.202 \pm 0.104$ & $0.155 \pm 0.019$ & $0.081 \pm 0.060$ & $0.048 \pm 0.039$ \\
TopoDSGC+Gprompt & $0.914 \pm 0.034$ & $0.779 \pm 0.037$ & $0.879 \pm 0.072$ & $0.825 \pm 0.146$ & $0.176 \pm 0.027$ & $0.100 \pm 0.006$ \\
TopoDSGC+All-in-one & \underline{$0.922 \pm 0.014$} & \underline{$0.780 \pm 0.056$} & $\mathbf{0.909 \pm 0.066}$ & $0.843 \pm 0.158$ & $0.168 \pm 0.035$ & $0.096 \pm 0.003$ \\
TopoDSGC+GPF & $\mathbf{0.938 \pm 0.017}$ & $0.737 \pm 0.111$ & \underline{$0.907 \pm 0.038$} & $0.853 \pm 0.169$ & $0.182 \pm 0.026$ & $0.101 \pm 0.002$ \\
TopoSigMaNet+GPPT & $-0.001 \pm 0.002$ & $0.008 \pm 0.004$ & $0.111 \pm 0.037$ & $0.231 \pm 0.093$ & $0.039 \pm 0.055$ & $0.008 \pm 0.007$ \\
TopoSigMaNet+Gprompt & $0.069 \pm 0.020$ & $0.122 \pm 0.044$ & $0.136 \pm 0.052$ & $0.216 \pm 0.079$ & $0.116 \pm 0.030$ & $0.053 \pm 0.004$ \\
TopoSigMaNet+GPF & $0.118 \pm 0.039$ & $0.134 \pm 0.026$ & $0.161 \pm 0.041$ & $0.283 \pm 0.049$ & $0.118 \pm 0.037$ & $0.042 \pm 0.017$ \\
TopoSigMaNet+All-in-one & $0.114 \pm 0.026$ & $0.123 \pm 0.024$ & $0.150 \pm 0.029$ & $0.301 \pm 0.048$ & $0.140 \pm 0.024$ & $0.042 \pm 0.017$ \\
\bottomrule
\end{tabular}%
}
\end{table}

\begin{table}[htbp]
\centering
\caption{Prompt method comparison without topological information (5-shot without topo).}
\label{tab:ablation_5shot_no_topo_aligned}
\resizebox{\textwidth}{!}{%
\begin{tabular}{lcccccc}
\toprule
\textbf{Method} & \textbf{SDSBM-1} & \textbf{SDSBM-2} & \textbf{SDSBM-3} & \textbf{SDSBM-4} & \textbf{Rainfall} & \textbf{SP1500} \\
\midrule
MSGNN+GPPT & $0.030 \pm 0.026$ & $0.066 \pm 0.024$ & $0.363 \pm 0.025$ & $0.158 \pm 0.025$ & $0.140 \pm 0.018$ & $0.082 \pm 0.012$ \\
MSGNN+Gprompt & $0.596 \pm 0.056$ & $0.622 \pm 0.033$ & $0.715 \pm 0.041$ & $0.855 \pm 0.166$ & $0.163 \pm 0.018$ & $0.110 \pm 0.005$ \\
MSGNN+GPF & $0.575 \pm 0.066$ & $0.594 \pm 0.055$ & $0.773 \pm 0.016$ & \underline{$0.954 \pm 0.045$} & $0.170 \pm 0.017$ & $0.107 \pm 0.006$ \\
MSGNN+All-in-one & $0.579 \pm 0.039$ & $0.645 \pm 0.035$ & $0.724 \pm 0.024$ & $\mathbf{0.960 \pm 0.034}$ & $0.143 \pm 0.041$ & \underline{$0.110 \pm 0.007$} \\
\midrule
SSSNET+GPPT & $0.055 \pm 0.016$ & $0.099 \pm 0.012$ & $0.379 \pm 0.065$ & $0.163 \pm 0.018$ & $0.121 \pm 0.043$ & $0.067 \pm 0.012$ \\
SSSNET+Gprompt & $0.693 \pm 0.035$ & $0.773 \pm 0.059$ & $0.844 \pm 0.037$ & $0.891 \pm 0.131$ & $0.217 \pm 0.039$ & $0.104 \pm 0.012$ \\
SSSNET+GPF & $0.734 \pm 0.056$ & $0.798 \pm 0.028$ & $0.858 \pm 0.058$ & $0.921 \pm 0.126$ & \underline{$0.229 \pm 0.015$} & $0.107 \pm 0.015$ \\
SSSNET+All-in-one & $0.665 \pm 0.035$ & $0.760 \pm 0.061$ & $0.833 \pm 0.043$ & $0.943 \pm 0.086$ & $\mathbf{0.240 \pm 0.011}$ & $\mathbf{0.125 \pm 0.024}$ \\
\midrule
DSGC+GPPT & $0.042 \pm 0.019$ & $0.110 \pm 0.029$ & $0.318 \pm 0.040$ & $0.232 \pm 0.136$ & $0.102 \pm 0.051$ & $0.053 \pm 0.037$ \\
DSGC+Gprompt & $\mathbf{0.947 \pm 0.018}$ & $\mathbf{0.849 \pm 0.026}$ & $0.926 \pm 0.039$ & $0.861 \pm 0.170$ & $0.180 \pm 0.015$ & $0.095 \pm 0.006$ \\
DSGC+All-in-one & $0.928 \pm 0.021$ & \underline{$0.840 \pm 0.034$} & $\mathbf{0.953 \pm 0.018}$ & $0.865 \pm 0.162$ & $0.188 \pm 0.023$ & $0.088 \pm 0.018$ \\
DSGC+GPF & \underline{$0.941 \pm 0.020$} & $0.814 \pm 0.016$ & \underline{$0.945 \pm 0.014$} & $0.861 \pm 0.170$ & $0.198 \pm 0.010$ & $0.098 \pm 0.007$ \\
\midrule
SigMaNet+GPPT & $-0.001 \pm 0.003$ & $0.004 \pm 0.009$ & $0.007 \pm 0.009$ & $0.146 \pm 0.181$ & $0.000 \pm 0.000$ & $0.003 \pm 0.007$ \\
SigMaNet+Gprompt & $0.143 \pm 0.082$ & $0.047 \pm 0.085$ & $0.195 \pm 0.162$ & $0.665 \pm 0.151$ & $0.095 \pm 0.129$ & $0.030 \pm 0.036$ \\
SigMaNet+GPF & $0.097 \pm 0.095$ & $0.101 \pm 0.123$ & $0.072 \pm 0.106$ & $0.662 \pm 0.103$ & $0.000 \pm 0.000$ & $0.039 \pm 0.038$ \\
SigMaNet+All-in-one & $0.186 \pm 0.122$ & $0.048 \pm 0.089$ & $0.249 \pm 0.233$ & $0.678 \pm 0.178$ & $0.071 \pm 0.092$ & $0.030 \pm 0.036$ \\
\bottomrule
\end{tabular}%
}
\end{table}

\begin{table}[htbp]
\centering
\caption{Ablation study on 5-shot learning: performance improvement by adding topological information (5-shot with Topo minus 5-shot without Topo). Positive values represent gains from adding topological information.}
\label{tab:ablation_5shot_topo_difference}
\resizebox{\textwidth}{!}{%
\begin{tabular}{lcccccc}
\toprule
\textbf{Method} & \textbf{SDSBM-1} & \textbf{SDSBM-2} & \textbf{SDSBM-3} & \textbf{SDSBM-4} & \textbf{Rainfall} & \textbf{SP1500} \\
\midrule
MSGNN+GPPT & $+0.006$ & $+0.002$ & $-0.014$ & $+0.002$ & $+0.003$ & $+0.017$ \\
MSGNN+Gprompt & $+0.007$ & $+0.001$ & $-0.024$ & $+0.002$ & $+0.014$ & $-0.002$ \\
MSGNN+GPF & $0.000$ & $-0.026$ & $-0.038$ & $+0.016$ & $+0.031$ & $+0.001$ \\
MSGNN+All-in-one & $+0.036$ & $-0.055$ & $-0.037$ & $-0.105$ & $+0.027$ & $-0.008$ \\
\midrule
SSSNET+GPPT & $+0.036$ & $+0.023$ & $+0.002$ & $-0.012$ & $+0.009$ & $0.000$ \\
SSSNET+Gprompt & $-0.029$ & $-0.001$ & $+0.011$ & $+0.022$ & $-0.006$ & $+0.010$ \\
SSSNET+GPF & $-0.025$ & $-0.013$ & $-0.022$ & $-0.040$ & $-0.012$ & $-0.011$ \\
SSSNET+All-in-one & $+0.015$ & $+0.015$ & $+0.003$ & $+0.011$ & $-0.017$ & $-0.011$ \\
\midrule
DSGC+GPPT & $-0.013$ & $-0.038$ & $-0.116$ & $-0.077$ & $-0.021$ & $-0.005$ \\
DSGC+Gprompt & $-0.033$ & $-0.070$ & $-0.047$ & $-0.036$ & $-0.004$ & $+0.005$ \\
DSGC+All-in-one & $-0.006$ & $-0.060$ & $-0.044$ & $-0.022$ & $-0.020$ & $+0.008$ \\
DSGC+GPF & $-0.003$ & $-0.077$ & $-0.038$ & $-0.008$ & $-0.016$ & $+0.003$ \\
\midrule
SigMaNet+GPPT & $0.000$ & $+0.004$ & $+0.104$ & $+0.085$ & $+0.039$ & $+0.005$ \\
SigMaNet+Gprompt & $-0.074$ & $+0.075$ & $-0.059$ & $-0.449$ & $+0.021$ & $+0.023$ \\
SigMaNet+GPF & $+0.021$ & $+0.033$ & $+0.089$ & $-0.379$ & $+0.118$ & $+0.003$ \\
SigMaNet+All-in-one & $-0.072$ & $+0.075$ & $-0.099$ & $-0.377$ & $+0.069$ & $+0.012$ \\
\bottomrule
\end{tabular}%
}
\end{table}

\newpage
\section{Implementation details}
\label{app_sec:implementation}

\subsection{Dataset details}
Table \ref{tab:data-statistics} summarizes the dataset statistics, including the number of nodes, number of ground-truth clusters, positive and negative edge counts, and average degree for the four synthetic SDSBM settings and the two real-world signed graph datasets, Rainfall and SP1500.

\begin{table}[h!]
\centering
\caption{Dataset summary statistics, where SDSBM results are averaged over five random seeds and we also report one standard deviation.}
\label{tab:data-statistics}
\begin{tabular}{lrrrrrrrr}
\toprule
Dataset &  $n$ & $K$ & $|\mathcal{E}^+|$ & $|\mathcal{E}^-|$ & Avg. degree \\
\midrule
SDSBM-1 & 1,000 & 3 & 25,426 $\pm$ 109 & 24,565 $\pm$ 76 & 49.99 $\pm$ 0.12 \\
SDSBM-2 & 1,000 & 4 & 21,949 $\pm$ 101 & 27,958 $\pm$ 128 & 49.91 $\pm$ 0.13 \\
SDSBM-3 & 1,000 & 4 & 21,514 $\pm$ 129 & 28,433 $\pm$ 163 & 49.95 $\pm$ 0.20 \\
SDSBM-4 & 1,000 & 3 & 25,913 $\pm$ 196 & 24,013 $\pm$ 144 & 49.93 $\pm$ 0.23 \\
Rainfall & 306 & 6 & 64,408 & 29,228 & 306.00 \\
SP1500 & 1,193 & 10 & 1,069,319 & 353,930 & 1193.00 \\
\bottomrule
\end{tabular}%
\end{table}

\subsection{Setup}
Experiments were conducted on one compute node with one Nvidia H100 GPU with driver version 580.95.05 and CUDA version 13.0, 20 Intel(R) Xeon(R) Platinum 8468 CPUs and $234$GB RAM.

\subsection{Hyperparameters}
\label{app_subsec:hyperparam}
All experiments use a unified training protocol to ensure fair comparison across methods. Table~\ref{tab:hyperparameters} details the hyperparameter settings. In particular:

\begin{table}[htb!]
\centering
\caption{Hyperparameter settings used across all experiments.}
\label{tab:hyperparameters}
\resizebox{\textwidth}{!}{%
\begin{tabular}{llll}
\toprule
\textbf{Category} & \textbf{Hyperparameter} & \textbf{Value} & \textbf{Applies to} \\
\midrule
\multirow{7}{*}{Architecture}
  & GNN hidden dim ($d$)               & 32                  & All GNNs \\
  & TPL output dim ($d_T$)             & 32                  & TopoSIGN, TopoDIG \\
  & SigMaNet hidden (complex channels) & 1                   & SigMaNet only \\
  & GNN layers                         & 2                   & All GNNs \\
  & Chebyshev order ($K$)              & 1                   & MSGNN, TopoSIGN, TopoDIG, GFMs \\
  & Magnetic charge ($q$)              & 0.25                & MSGNN, TopoSIGN, TopoDIG, GFMs \\
  & Laplacian normalization            & symmetric           & All MSConv-based models \\
    & Dropout                            & 0.5 (0.0 for SAMGPT)& All \\
\midrule
\multirow{6}{*}{Node clustering (baselines)}
  & Optimizer                          & Adam                & All \\
  & Epochs                             & 1{,}000             & All \\
  & Learning rate           & $0.01$ & All \\
  & Weight decay                       & $5\times10^{-4}$    & All \\
  & Early stopping patience            & 400 epochs          & All \\
\midrule
\multirow{5}{*}{GFM pre-training}
  & Optimizer                          & Adam                & All GFMs \\
  & Epochs                             & 1{,}000             & All GFMs \\
  & Learning rate (fixed)              & $10^{-2}$           & All GFMs \\
  & Early stopping patience            & 400 epochs          & All GFMs \\
  & Link-sign batch size               & 64                  & All GFMs \\
\midrule
\multirow{4}{*}{GFM fine-tuning}
  & Optimizer                          & Adam                & All GFMs \\
  & Epochs                             & 100                 & All GFMs \\
  & Early stopping patience            & 40 epochs           & All GFMs \\
  & Learning rate (searched)           & $\{10^{-2},\,5{\times}10^{-3},\,10^{-3},\,5{\times}10^{-4},\,10^{-4}\}$ & selected by val ARI \\
\midrule
\multirow{5}{*}{PI computation (TDA)}
  & $k$-hop neighborhood               & 2                   & TopoSIGN, TopoDIG \\
  & Filtration max scale               & 10.0                & TopoSIGN, TopoDIG \\
  & Pixel size                         & 1.0                 & TopoSIGN, TopoDIG \\
  & PI grid / feature dim              & $10\times10 = 100$  & TopoSIGN, TopoDIG \\
  & Top-$k$ landmarks                  & 80                  & TopoSIGN, TopoDIG \\
\midrule
\multirow{3}{*}{Data splits}
  & Node train set                      & 10\%                & All \\
  & Node validation set                & 10\%                & All \\
  & Node test set                & 80\%                & All \\
\midrule
\multirow{2}{*}{Evaluation}
  & Training seeds                     & 5 $(\{0,10,20,30,40\})$ & All \\
  & Metric                             & ARI (mean $\pm$ std)    & All \\
\bottomrule
\end{tabular}%
}
\end{table}

\paragraph{Optimizer and training duration.} All models are trained with the Adam optimizer (learning rate 0.01 by default, weight decay $5\times 10^{-4}$, dropout 0.5) for up to 1,000 epochs with early stopping (patience 400 epochs). Pre-training fine-tuning stages use a shorter budget of 100 epochs and patience 40. All metrics are reported as the mean ± standard deviation over 5 independent training seeds (0, 10, 20, 30, 40).

\paragraph{Data splits.} The graph is split once per dataset instance using deterministic seeds. For node clustering, each graph is partitioned into 10\% train (seed) nodes (labeled for NLL loss), 10\% validation nodes (for ARI-based early stopping), and 80\% test nodes (on which the final ARI is reported). 

\paragraph{Hyperparameter selection.} The only tuned hyperparameter is the learning rate during finetuning, searched over $\{10^{-2},\,5{\times}10^{-3},\,10^{-3},\,5{\times}10^{-4},\,10^{-4}\}$. All other hyperparameters are fixed across all methods and datasets: hidden dimension 32, topological hidden dimension 32, Chebyshev order K=1, magnetic charge q=0.25, 2 MSConv layers, normalization='sym', and persistence-image pixel size 1.0. The best learning rate is selected per method and dataset per seed by validation ARI. Multiclass node objectives use finite NLL loss on log-softmax outputs.
\paragraph{Balanced pre-training mini-batches.}
During pre-training, the GFM encoder is trained on the observed signed directed graph
$\mathcal{G} = (\mathcal{V}, \mathcal{E}^+, \mathcal{E}^-)$,
where $\mathcal{E}^+$ and $\mathcal{E}^-$ denote the positive and negative edge sets.
We use all observed edges as the pre-training pool; no separate link split is held out,
because pre-training is unsupervised and is fully decoupled from the node-clustering
evaluation splits.

At each pre-training epoch we construct a mini-batch of size $B = 64$ by
\emph{stratified sign sampling}: we draw $B/2$ edges uniformly at random from
$\mathcal{E}^+$ and $B/2$ from $\mathcal{E}^-$, yielding a balanced batch
$\mathcal{B} = \mathcal{B}^+ \cup \mathcal{B}^-$ with $|\mathcal{B}^+| = |\mathcal{B}^-| = B/2$.
If one sign class is absent (e.g.\ for unsigned graphs), we revert to uniform random
sampling over all edges.
This balancing is important because real-world signed networks are often heavily skewed
toward positive edges~\cite{leskovec2010predicting}, and an unbalanced batch would bias
the binary cross-entropy loss of the sign-prediction (SP) objective.
Given node embeddings $\mathbf{Z}$ produced by the frozen backbone, an MLP decoder scores each edge $(u,v)\in\mathcal{B}$ as positive or negative via
    $\hat{y}_{uv} = \sigma\!\left(\text{MLP}([\mathbf{z}_u \|\mathbf{z}_v])\right)$,
    optimized with binary cross-entropy.
    The balanced batch ensures equal gradient contributions from both sign classes.

The structured auxiliary objectives used in Table~\ref{tab:ablation_pretrain} are derived from the same balanced
batch $\mathcal{B}$, so their class distributions are naturally balanced as well:

\begin{itemize}[leftmargin=*,noitemsep]
  \item \textbf{SP} (sign prediction).
    Given node embeddings $\mathbf{Z}$ produced by the frozen backbone, an MLP decoder
    scores each edge $(u,v)\in\mathcal{B}$ as positive or negative via
    $\hat{y}_{uv} = \sigma\!\left(\text{MLP}([\mathbf{z}_u \|\mathbf{z}_v])\right)$,
    optimized with binary cross-entropy.
    The balanced batch ensures equal gradient contributions from both sign classes.

  \item \textbf{DP} (direction prediction).
    The reversed batch $\mathcal{B}_{\text{rev}} = \{(v,u)\mid(u,v)\in\mathcal{B}\}$
    is constructed by flipping the source and destination indices.
    The combined pool $\mathcal{B} \cup \mathcal{B}_{\text{rev}}$ (size $2B$) is labeled
    $1$ for forward and $0$ for reversed edges; the 1:1 ratio is preserved exactly.

  \item \textbf{3C} (three-class sign and existence prediction).
    Non-edges $\mathcal{N}$ of size $|\mathcal{B}|$ are sampled via sparse negative
    sampling~\citep{hamilton2017inductive} from node pairs absent in
    $\mathcal{E}^+\cup\mathcal{E}^-$.
    Labels are $0$ (positive forward), $1$ (negative forward), and $2$ (non-edge),
    with $B/2$ edges per observed class and $B$ non-edges, giving a
    $\tfrac{1}{4}{:}\tfrac{1}{4}{:}\tfrac{1}{2}$ class ratio.

  \item \textbf{4C} (four-class directed sign prediction).
    The reversed batch is concatenated to form
    $\mathcal{B} \cup \mathcal{B}_{\text{rev}}$ (size $2B$).
    Labels encode all four sign$\times$direction combinations
    (positive forward, negative forward, positive reversed, negative reversed),
    each of size $B/2$, yielding a perfectly balanced four-class distribution.

  \item \textbf{5C} (five-class directed sign and existence prediction).
    Non-edges $\mathcal{N}$ (size $B$) are added to the 4C pool, giving five classes
    of sizes $B/2$, $B/2$, $B/2$, $B/2$, $B$; the four observed classes remain balanced.
\end{itemize}
The model is pre-trained for up to $1{,}000$ epochs with Adam (learning rate $10^{-2}$,
weight decay $5\times 10^{-4}$) and early stopping with patience $400$ on the training
loss. Fine-tuning uses a separate learning rate selected from
$\{10^{-2}, 5\times10^{-3}, 10^{-3}, 5\times10^{-4}, 10^{-4}\}$ via validation ARI,
with the pre-trained weights fixed as the starting point.

\paragraph{Evaluation metric.} We report Adjusted Rand Index (ARI) on the test split (~80\% of nodes). ARI equals 1 for perfect agreement with ground-truth clusters and 0 for random clustering.

\subsection{Runtime and Computational Cost}
\label{app_subsec:runtime}
To provide a transparent breakdown of the computational cost, we explicitly report the runtime for preprocessing, model training, and the whole running time:
\begin{itemize}[leftmargin=1.5em, itemsep=3pt, topsep=2pt]
    \item \textbf{Training and Evaluation Runtime (After Preprocessing):} Table~\ref{tab:runtime_main_compare_gfms} reports the runtime (pre-training, prompt-tuning, and evaluation per run/seed) for graph prompt learning methods. Table~\ref{tab:runtime_main_results_other_signed_GNNs} reports the training and evaluation runtime for standard signed GNNs and signed GNNs plus the topological branch in semi-supervised clustering.
    \item \textbf{Preprocessing Time:} Persistent Image (PI) extraction is computed only once as a preprocessing step prior to model training. The exact preprocessing time per dataset is detailed in Table~\ref{tab:runtime_pi_computation}. Because persistence diagrams are extracted independently for each node's local ego-network, this step is easily parallelizable across CPU cores.
    \item \textbf{Whole Running Time:} The total running time for our methods consists of the one-time preprocessing time in Table~\ref{tab:runtime_pi_computation} plus the training and evaluation time in Table~\ref{tab:runtime_main_compare_gfms} (or Table~\ref{tab:runtime_main_results_other_signed_GNNs}). Once topological features are computed and stored, they do not add any computational cost during repeated training, tuning, or evaluation.
\end{itemize}

The empirical results reveal a clear and practical trade-off. As detailed in Table \ref{tab:runtime_pi_computation}, the initial feature extraction takes roughly 2 to 32 minutes for sparse and moderately dense graphs (e.g., Rainfall, SDSBMs), though it scales up significantly on dense, complete graphs like SP1500. However, because each node's ego-network is processed independently, this upfront whole-time bottleneck is highly parallelizable across standard CPU cores. 

More importantly, this decoupled design successfully shifts the computational burden away from the iterative model training and evaluation loops. As shown in Tables \ref{tab:runtime_main_compare_gfms} and \ref{tab:runtime_main_results_other_signed_GNNs}, once the topological summaries are precomputed, ingesting them introduces only a moderate overhead during training. For instance, incorporating the topological branch typically extends the training time by less than a factor of two compared to the bare signed GNN backbones, keeping the training-only cost firmly on the order of a few minutes per run. Consequently, while the whole running time on a newly seen dataset must account for the initial preprocessing phase, any subsequent operations only incur the highly competitive training-only runtime. We believe this demonstrates that TopoSIGN achieves performance gains while remaining computationally practical.

\begin{table}[htbp]
\centering
\caption{Average runtime (seconds) per run across different seeds for graph prompt learning methods (corresponding to Table~\ref{tab:main_compare_gfms}), covering pre-training, prompt-tuning, and evaluation \textbf{after} preprocessing. The shortest time is marked in \textbf{bold} while the second shortest time is \underline{underlined}.}
\label{tab:runtime_main_compare_gfms}
\begin{tabular}{lcccccc}
\toprule
\textbf{Method} & \textbf{SDSBM-1} & \textbf{SDSBM-2} & \textbf{SDSBM-3} & \textbf{SDSBM-4} & \textbf{Rainfall} & \textbf{SP1500} \\
\midrule
GPPT & $95 \pm 11$ & $109 \pm 15$ & $129 \pm 13$ & $118 \pm 16$ & $168 \pm 12$ & $285 \pm 47$ \\
Gprompt & $68 \pm 16$ & $73 \pm 26$ & $93 \pm 7$ & \underline{$77 \pm 10$} & $80 \pm 17$ & $126 \pm 20$ \\
GPF & $69 \pm 7$ & $\mathbf{69 \pm 18}$ & $\mathbf{81 \pm 19}$ & $84 \pm 17$ & $98 \pm 10$ & $128 \pm 26$ \\
All-in-one & \underline{$66 \pm 13$} & $70 \pm 12$ & $95 \pm 5$ & $\mathbf{74 \pm 14}$ & \underline{$80 \pm 14$} & \underline{$123 \pm 20$} \\
SAMGPT & $\mathbf{64 \pm 14}$ & \underline{$70 \pm 16$} & \underline{$85 \pm 16$} & $83 \pm 13$ & $\mathbf{77 \pm 15}$ & $\mathbf{97 \pm 10}$ \\
TopoDIG & $141 \pm 12$ & $203 \pm 26$ & $177 \pm 19$ & $179 \pm 15$ & $218 \pm 28$ & $304 \pm 85$ \\
TopoMSGNN & $324 \pm 36$ & $399 \pm 42$ & $304 \pm 32$ & $227 \pm 17$ & $270 \pm 56$ & $386 \pm 102$ \\
TopoSSSNET & $879 \pm 104$ & $656 \pm 84$ & $670 \pm 56$ & $629 \pm 137$ & $1003 \pm 225$ & $1380 \pm 278$ \\
\bottomrule
\end{tabular}
\end{table}

\begin{table}[htbp]
\centering
\caption{Average training and evaluation runtime (seconds) per run across different seeds for signed GNNs and signed GNNs plus the topological branch (corresponding to Table~\ref{tab:main_results_other_signed_GNNs}), excluding one-time preprocessing. The shortest time is marked in \textbf{bold} while the second shortest time is \underline{underlined}.}
\label{tab:runtime_main_results_other_signed_GNNs}
\begin{tabular}{lcccccc}
\toprule
\textbf{Method} & \textbf{SDSBM-1} & \textbf{SDSBM-2} & \textbf{SDSBM-3} & \textbf{SDSBM-4} & \textbf{Rainfall} & \textbf{SP1500} \\
\midrule
DSGC & $\mathbf{54 \pm 3}$ & $\mathbf{56 \pm 5}$ & $\mathbf{56 \pm 2}$ & $\mathbf{61 \pm 17}$ & $\mathbf{56 \pm 2}$ & $\mathbf{82 \pm 7}$ \\
SigMaNet & $140 \pm 47$ & $131 \pm 38$ & \underline{$90 \pm 8$} & $125 \pm 45$ & $135 \pm 45$ & $188 \pm 45$ \\
MSGNN & $94 \pm 32$ & \underline{$78 \pm 11$} & $109 \pm 29$ & \underline{$64 \pm 2$} & \underline{$88 \pm 18$} & $224 \pm 73$ \\
SSSNET & $100 \pm 4$ & $99 \pm 1$ & $106 \pm 18$ & $99 \pm 1$ & $130 \pm 56$ & \underline{$170 \pm 22$} \\
\midrule
DSGC+Topo & \underline{$94 \pm 32$} & $120 \pm 33$ & $127 \pm 23$ & $81 \pm 14$ & $150 \pm 26$ & $201 \pm 53$ \\
SigMaNet+Topo & $158 \pm 43$ & $182 \pm 32$ & $162 \pm 37$ & $165 \pm 47$ & $130 \pm 36$ & $223 \pm 51$ \\
MSGNN+Topo & $148 \pm 34$ & $147 \pm 36$ & $134 \pm 27$ & $91 \pm 13$ & $185 \pm 10$ & $298 \pm 43$ \\
SSSNET+Topo & $166 \pm 28$ & $163 \pm 29$ & $195 \pm 35$ & $146 \pm 17$ & $264 \pm 44$ & $409 \pm 78$ \\
\bottomrule
\end{tabular}
\end{table}

\begin{table}[htbp]
\centering
\caption{One-time Persistent Image (PI) preprocessing time (in seconds) across datasets. This step is executed once prior to training and can be parallelized across CPU cores.}
\label{tab:runtime_pi_computation}
\begin{tabular}{lcccccc}
\toprule
\textbf{Dataset} & \textbf{SDSBM-1} & \textbf{SDSBM-2} & \textbf{SDSBM-3} & \textbf{SDSBM-4} & \textbf{Rainfall} & \textbf{SP1500} \\
\midrule
PI computation (s) & 1917.12 & 1924.32 & 1928.89 & 1904.11 & 145.67 & 9435.03 \\
\bottomrule
\end{tabular}
\end{table}

\subsection{Parameter Sensitivity Analysis}

\subsubsection{Sensitivity Analysis on Ego-Network Radius ($k$-hop)}
\label{app_subsec:sensitivity_khop}
In our main experiments (Table~\ref{tab:main_compare_gfms} and Table~\ref{tab:main_results_other_signed_GNNs}), the topological features are extracted from 2-hop ego-networks ($k=2$) with a pixel size of $1.0$. To investigate the sensitivity and robustness of TopoSIGN with respect to the ego-network neighborhood radius, we evaluate our framework under an expanded 3-hop context ($k=3$, $\text{pixel\_size}=1.0$). Table~\ref{tab:sensitivity_khop_gfms} and Table~\ref{tab:sensitivity_khop_signed_gnns} present the downstream node clustering ARI for graph prompt learning and semi-supervised signed GNNs, respectively. This comparison isolates the impact of higher-order multi-hop connectivity patterns against localized neighborhood topology.

The empirical results reveal a nuanced but consistent trend. In the prompt learning setting (Table \ref{tab:sensitivity_khop_gfms}), increasing the radius to $k = 3$ yields highly stable performance with only marginal fluctuations compared to $k = 2$. In the semi-supervised setting (Table \ref{tab:sensitivity_khop_signed_gnns}), expanding the receptive field to $k = 3$ actually brings noticeable ARI improvements on several synthetic datasets (e.g., DSGC+Topo and SigMaNet+Topo on SDSBM-1 to 3), while maintaining comparable results on real-world networks like Rainfall and SP1500.

Despite the performance gains of $k = 3$ on certain baselines, we set $k = 2$ as the default in our main experiments to strike an optimal balance between topological discriminability and computational efficiency. Because a 2-hop ego-network is mathematically sufficient to capture the fundamental building blocks of structural balance (i.e., signed triangles and local cycles), it provides strong structural regularization without incurring the larger offline preprocessing cost required to extract persistence diagrams from 3-hop dense subgraphs.

\begin{table}[htbp]
\centering
\caption{Pre-training and prompt learning performance comparison in terms of downstream node clustering ARI across different ego-network radii ($k$-hop). We report the mean ARI among five runs plus/minus one standard deviation.}

\label{tab:sensitivity_khop_gfms}
\resizebox{\textwidth}{!}{%
\begin{tabular}{llcccccc}
\toprule
\textbf{$k$-hop} & \textbf{Method} & \textbf{SDSBM-1} & \textbf{SDSBM-2} & \textbf{SDSBM-3} & \textbf{SDSBM-4} & \textbf{Rainfall} & \textbf{SP1500} \\
\midrule
\multirow{2}{*}{$k=2$} & TopoMSGNN & $0.032 \pm 0.030$ & $0.107 \pm 0.009$ & $0.372 \pm 0.042$ & $0.164 \pm 0.025$ & $0.075 \pm 0.061$ & $0.000 \pm 0.000$ \\
                       & TopoSSSNET & $0.076 \pm 0.031$ & $0.125 \pm 0.011$ & $0.379 \pm 0.036$ & $0.157 \pm 0.023$ & $0.062 \pm 0.057$ & $0.000 \pm 0.000$ \\
\midrule
\multirow{2}{*}{$k=3$} & TopoMSGNN & $0.026 \pm 0.033$ & $0.090 \pm 0.021$ & $0.382 \pm 0.049$ & $0.167 \pm 0.017$ & $0.069 \pm 0.057$ & $0.009 \pm 0.018$ \\
                       & TopoSSSNET& $0.087 \pm 0.042$ & $0.121 \pm 0.018$ & $0.373 \pm 0.023$ & $0.165 \pm 0.009$ & $0.049 \pm 0.060$ & $0.000 \pm 0.000$ \\
\bottomrule
\end{tabular}%
}
\end{table}
\vspace{-5mm}
\begin{table}[htbp]
\centering
\caption{Semi-supervised node clustering performance (ARI) on real-world and synthetic signed datasets across different ego-network radii ($k$-hop). We report the mean ARI among five runs plus/minus one standard deviation.}
\label{tab:sensitivity_khop_signed_gnns}
\resizebox{\textwidth}{!}{%
\begin{tabular}{llcccccc}
\toprule
\textbf{$k$-hop} & \textbf{Method} & \textbf{SDSBM-1} & \textbf{SDSBM-2} & \textbf{SDSBM-3} & \textbf{SDSBM-4} & \textbf{Rainfall} & \textbf{SP1500} \\
\midrule
\multirow{4}{*}{$k=2$} & DSGC+Topo     & $0.762 \pm 0.133$ & $0.777 \pm 0.160$ & $0.911 \pm 0.083$ & $0.989 \pm 0.009$ & $0.472 \pm 0.044$ & $0.156 \pm 0.024$ \\
                       & SigMaNet+Topo & $0.169 \pm 0.015$ & $0.248 \pm 0.062$ & $0.284 \pm 0.034$ & $0.458 \pm 0.234$ & $0.174 \pm 0.042$ & $0.072 \pm 0.006$ \\
                       & MSGNN+Topo    & $0.684 \pm 0.022$ & $0.559 \pm 0.105$ & $0.674 \pm 0.054$ & $0.974 \pm 0.022$ & $0.435 \pm 0.084$ & $0.170 \pm 0.013$ \\
                       & SSSNET+Topo   & $0.923 \pm 0.030$ & $0.923 \pm 0.029$ & $0.937 \pm 0.047$ & $0.967 \pm 0.028$ & $0.481 \pm 0.034$ & $0.298 \pm 0.041$ \\
\midrule
\multirow{4}{*}{$k=3$} & DSGC+Topo     & $0.966 \pm 0.021$ & $0.905 \pm 0.024$ & $0.960 \pm 0.013$ & $0.958 \pm 0.071$ & $0.491 \pm 0.058$ & $0.157 \pm 0.036$ \\
                       & SigMaNet+Topo & $0.311 \pm 0.050$ & $0.352 \pm 0.137$ & $0.420 \pm 0.142$ & $0.609 \pm 0.126$ & $0.146 \pm 0.145$ & $0.064 \pm 0.032$ \\
                       & MSGNN+Topo    & $0.693 \pm 0.029$ & $0.671 \pm 0.011$ & $0.764 \pm 0.024$ & $0.990 \pm 0.009$ & $0.433 \pm 0.085$ & $0.169 \pm 0.014$ \\
                       & SSSNET+Topo   & $0.955 \pm 0.034$ & $0.940 \pm 0.014$ & $0.961 \pm 0.018$ & $1.000 \pm 0.000$ & $0.511 \pm 0.109$ & $0.295 \pm 0.072$ \\
\bottomrule
\end{tabular}%
}
\end{table}
\vspace{-5mm}
\begin{figure}[htbp]
    \centering
    \begin{subfigure}[b]{0.49\textwidth}
        \centering
        \includegraphics[width=\linewidth]{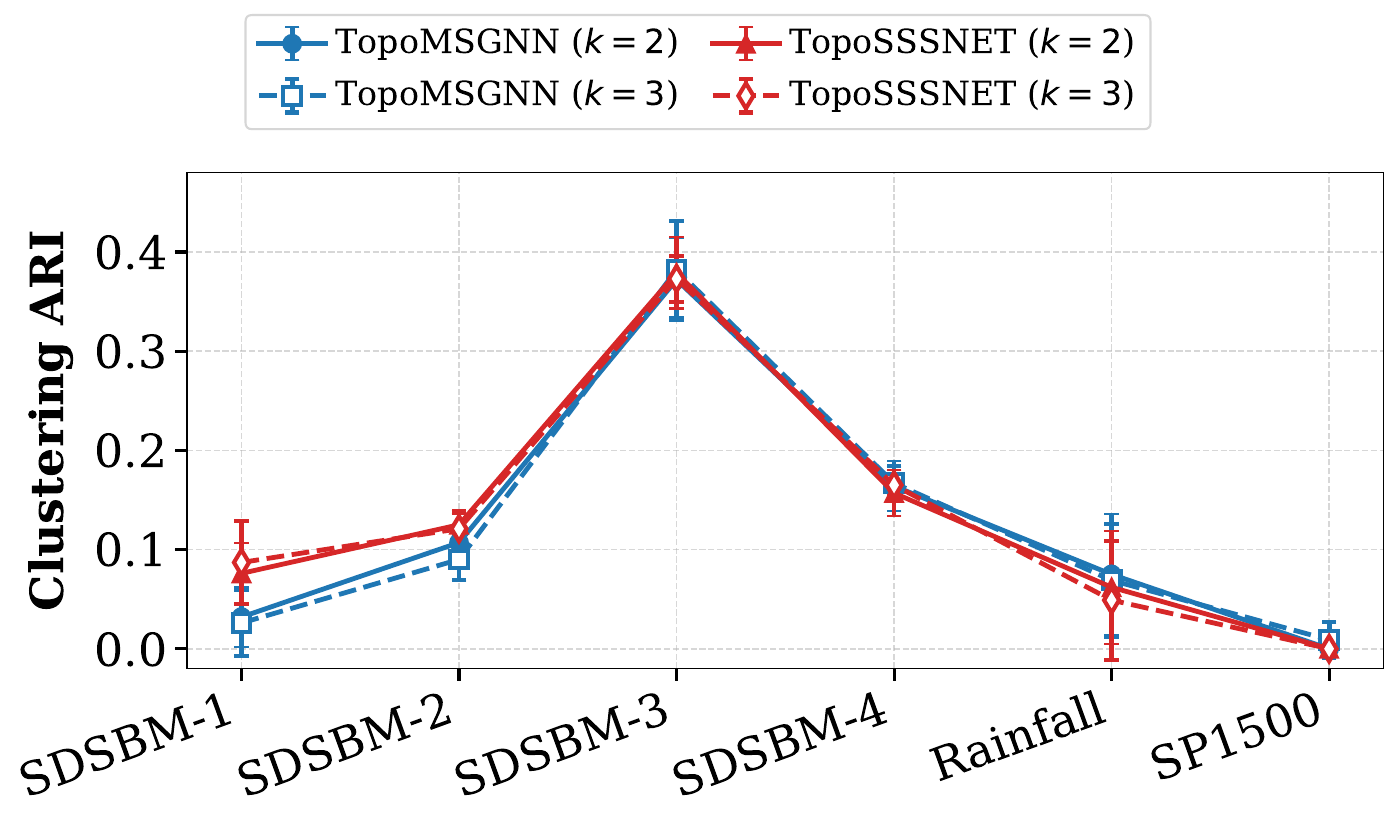}
        \caption{Graph Prompt Learning}
        \label{fig:sub_prompt_gfms}
    \end{subfigure}
    \hfill
    \begin{subfigure}[b]{0.49\textwidth}
        \centering
        \includegraphics[width=\linewidth]{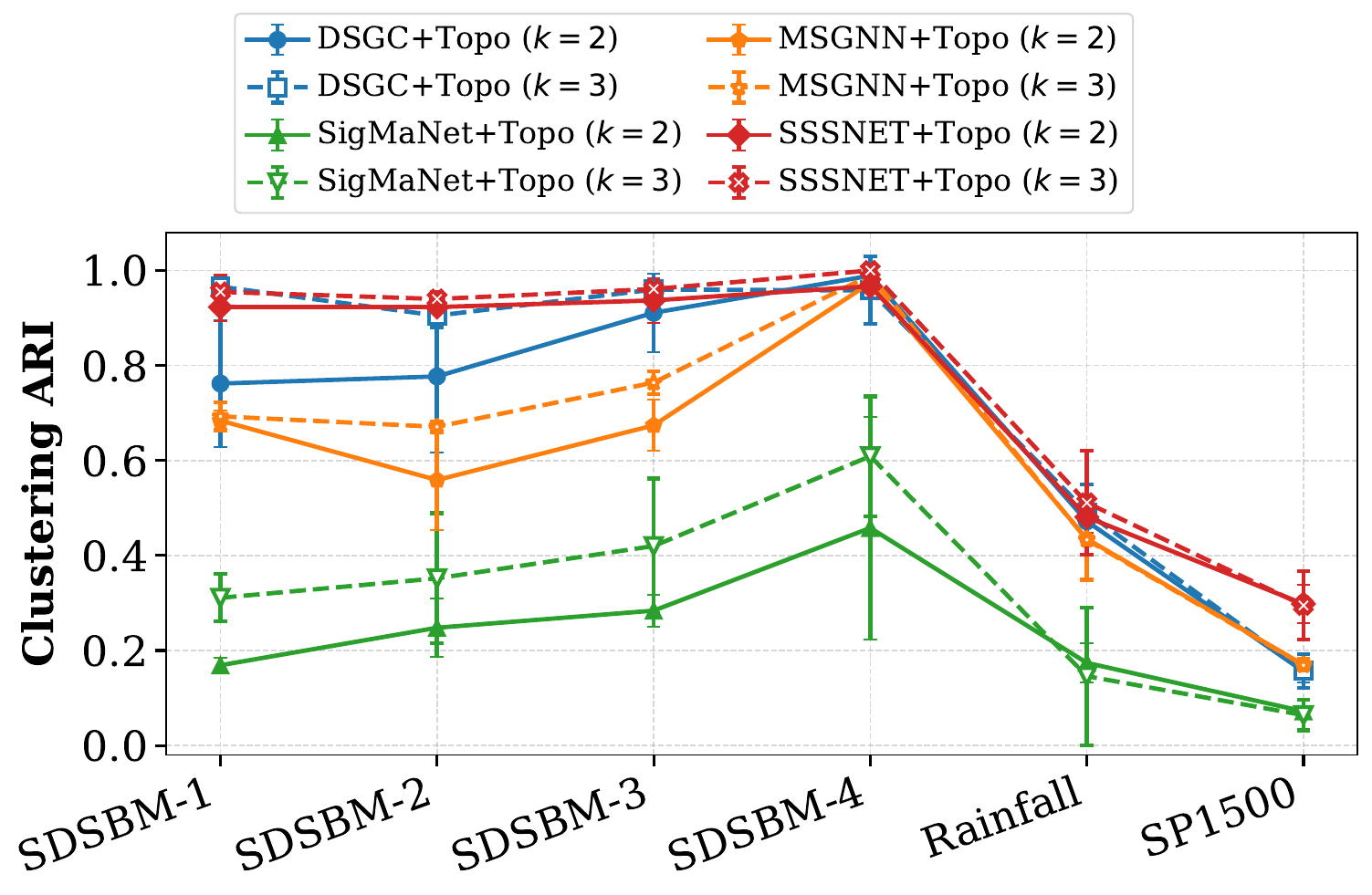}
        \caption{Semi-Supervised Signed GNNs}
        \label{fig:sub_signed_gnns}
    \end{subfigure}
    \caption{Downstream node clustering ARI sensitivity comparison across ego-network extraction radii ($k=2$ vs. $k=3$) over diverse datasets. Solid lines represent $k=2$ and dashed lines represent $k=3$.}
    \label{fig:sensitivity_khop_lines}
\end{figure} 
\vspace{-5mm}
\subsubsection{Sensitivity Analysis on Persistence Image Resolution (Pixel Size)}
\label{app_subsec:sensitivity_pixel_size}
The discretization resolution of persistence images directly determines the granularity of the vectorized topological representations. To evaluate model sensitivity to this hyperparameter, we conduct experiments across varying grid resolutions with $\text{pixel\_size} \in \{0.1, 1.0, 5.0\}$ while keeping the default 2-hop neighborhood ($k=2$). Table~\ref{tab:sensitivity_pixel_gfms} and Table~\ref{tab:sensitivity_pixel_signed_gnns} summarize the node clustering performance across pre-training and semi-supervised settings. These results demonstrate the stability of TopoSIGN across a reasonable range of topological feature discretizations and confirm that our framework does not require meticulous resolution tuning to achieve competitive performance.

The empirical results reveal that while the framework functions across different resolutions, grid granularity does meaningfully impact downstream accuracy. In the prompt learning setting (Table~\ref{tab:sensitivity_pixel_gfms}), performance remains relatively stable across resolutions with minor fluctuations. However, in the semi-supervised setting (Table~\ref{tab:sensitivity_pixel_signed_gnns}), deviating from the default resolution ($\text{pixel\_size} = 1.0$) noticeably degrades performance on several datasets (e.g., MSGNN+Topo and SSSNET+Topo on SDSBM-3, or MSGNN+Topo on SDSBM-4 under $\text{size}=5.0$).

This performance behavior aligns with the inherent mechanics of persistence images and reflects a trade-off in resolution granularity. A balanced grid ($\text{pixel\_size} = 1.0$) acts as an effective topological filter, aggregating nearby persistence points to produce dense, robust structural representations. Conversely, an overly fine grid ($\text{pixel\_size} = 0.1$) produces highly sparse feature matrices susceptible to localized topological noise, while an overly coarse grid ($\text{pixel\_size} = 5.0$) over-smoothes the topological landscape, sacrificing discriminative fine-grained structural features.
\begin{figure}[htbp]
    \centering
    \begin{subfigure}[b]{\linewidth}
        \centering
        \includegraphics[width=0.85\linewidth]{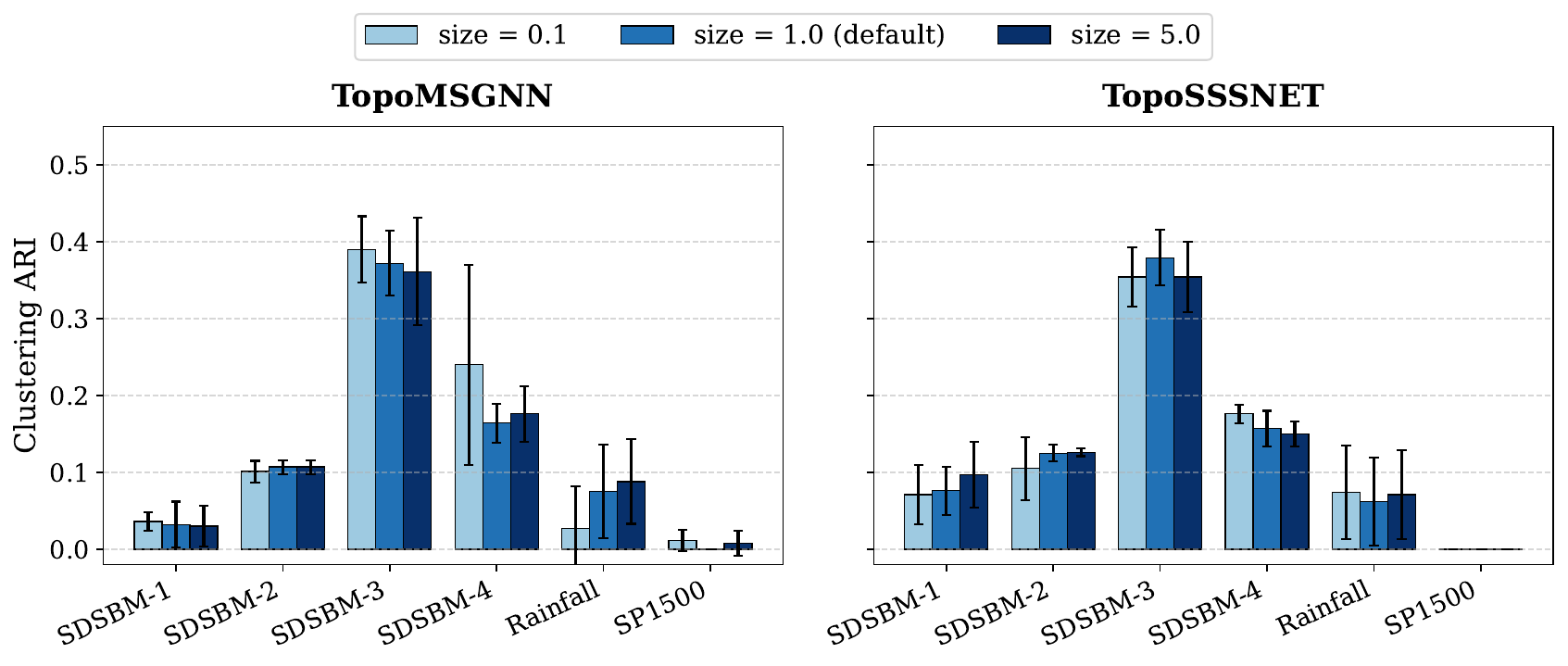}
        \caption{Graph Prompt Learning}
        \label{fig:sub_pixel_prompt_gfms}
    \end{subfigure}
    \vspace{0.8em} 
    \begin{subfigure}[b]{\linewidth}
        \centering
        \includegraphics[width=0.85\linewidth]{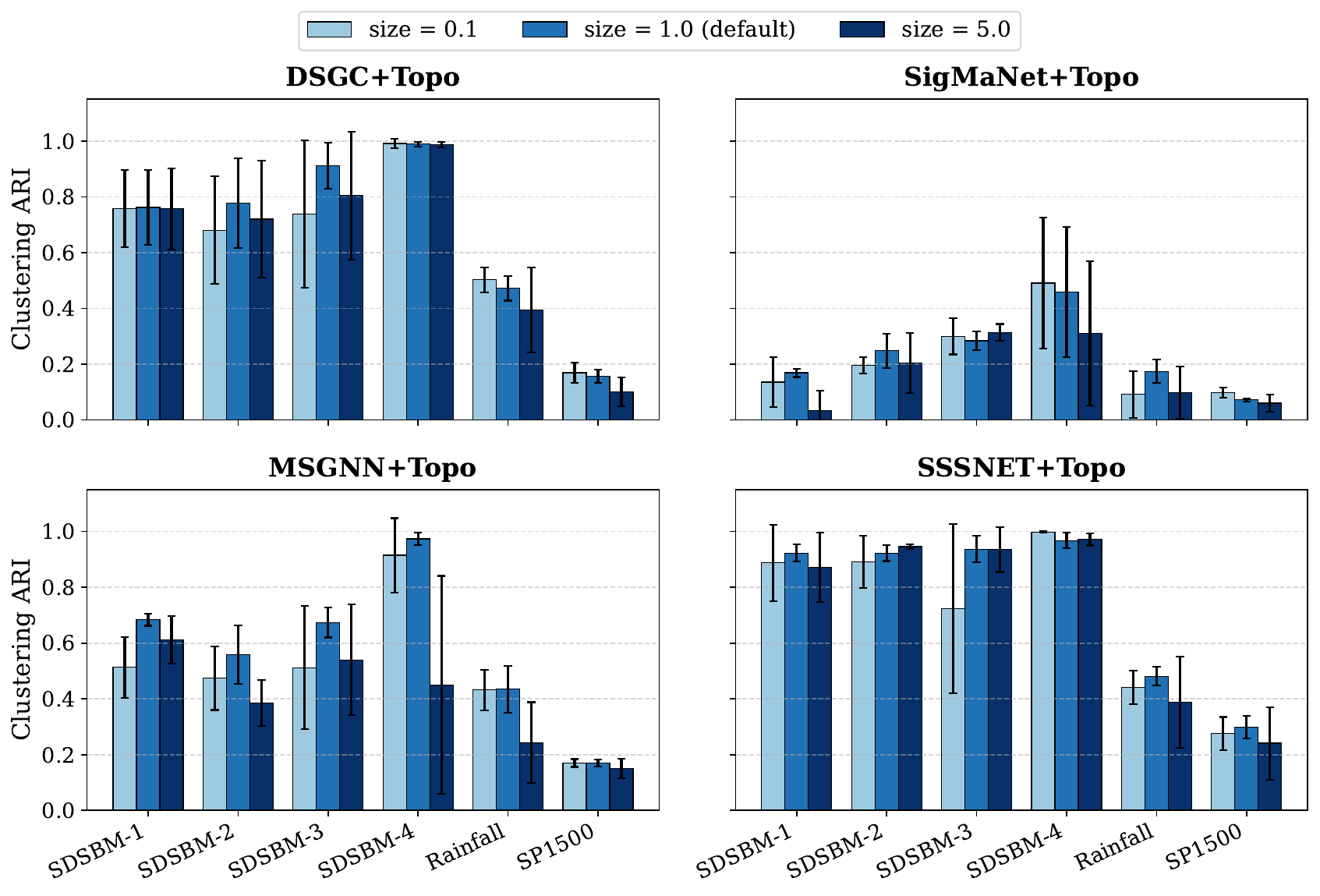}
        \caption{Semi-Supervised Signed GNNs}
        \label{fig:sub_pixel_signed_gnns}
    \end{subfigure}
    \vspace{-5mm}
    \caption{Downstream node clustering ARI sensitivity comparison across persistence image discretization resolutions ($\text{pixel\_size} \in \{0.1, 1.0, 5.0\}$) over diverse datasets. Bars from lighter to darker shades denote $\text{pixel\_size}=0.1$, $1.0$ (default), and $5.0$, respectively.}
    \label{fig:sensitivity_pixel_bars}
\end{figure}

\begin{table}[htbp]
\centering
\caption{Pre-training and prompt learning performance comparison in terms of downstream node clustering ARI across different persistence image pixel sizes. We report the mean ARI among five runs plus/minus one standard deviation.}
\label{tab:sensitivity_pixel_gfms}
\resizebox{\textwidth}{!}{%
\begin{tabular}{llcccccc}
\toprule
\textbf{Pixel Size} & \textbf{Method} & \textbf{SDSBM-1} & \textbf{SDSBM-2} & \textbf{SDSBM-3} & \textbf{SDSBM-4} & \textbf{Rainfall} & \textbf{SP1500} \\
\midrule
\multirow{2}{*}{$\text{size}=5.0$} & TopoMSGNN & $0.030 \pm 0.027$ & $0.107 \pm 0.009$ & $0.361 \pm 0.070$ & $0.176 \pm 0.036$ & $0.088 \pm 0.055$ & $0.008 \pm 0.016$ \\

                                   & TopoSSSNET & $0.097 \pm 0.043$ & $0.126 \pm 0.005$ & $0.354 \pm 0.046$ & $0.150 \pm 0.016$ & $0.071 \pm 0.058$ & $0.000 \pm 0.000$ \\
\midrule
\multirow{2}{*}{$\text{size}=1.0$} & TopoMSGNN  & $0.032 \pm 0.030$ & $0.107 \pm 0.009$ & $0.372 \pm 0.042$ & $0.164 \pm 0.025$ & $0.075 \pm 0.061$ & $0.000 \pm 0.000$ \\
                                   & TopoSSSNET& $0.076 \pm 0.031$ & $0.125 \pm 0.011$ & $0.379 \pm 0.036$ & $0.157 \pm 0.023$ & $0.062 \pm 0.057$ & $0.000 \pm 0.000$ \\
\midrule
\multirow{2}{*}{$\text{size}=0.1$} & TopoMSGNN  & $0.036 \pm 0.012$ & $0.101 \pm 0.014$ & $0.390 \pm 0.043$ & $0.240 \pm 0.130$ & $0.027 \pm 0.055$ & $0.011 \pm 0.014$ \\
                                   & TopoSSSNET & $0.071 \pm 0.039$ & $0.105 \pm 0.041$ & $0.354 \pm 0.039$ & $0.176 \pm 0.012$ & $0.074 \pm 0.061$ & $0.000 \pm 0.000$ \\
\bottomrule
\end{tabular}%
}
\end{table}

\begin{table}[htbp]
\centering
\caption{Semi-supervised node clustering performance (ARI) on real-world and synthetic signed datasets across different persistence image pixel sizes. We report the mean ARI among five runs plus/minus one standard deviation.}
\label{tab:sensitivity_pixel_signed_gnns}
\resizebox{\textwidth}{!}{%
\begin{tabular}{llcccccc}
\toprule
\textbf{Pixel Size} & \textbf{Method} & \textbf{SDSBM-1} & \textbf{SDSBM-2} & \textbf{SDSBM-3} & \textbf{SDSBM-4} & \textbf{Rainfall} & \textbf{SP1500} \\
\midrule
\multirow{4}{*}{$\text{size}=5.0$} & DSGC+Topo & $0.757 \pm 0.145$ & $0.720 \pm 0.210$ & $0.805 \pm 0.229$ & $0.988 \pm 0.010$ & $0.395 \pm 0.152$ & $0.101 \pm 0.052$ \\
                                   & SigMaNet+Topo & $0.034 \pm 0.070$ & $0.205 \pm 0.108$ & $0.314 \pm 0.030$ & $0.311 \pm 0.259$ & $0.098 \pm 0.094$ & $0.060 \pm 0.030$ \\
                                   & MSGNN+Topo & $0.611 \pm 0.085$ & $0.386 \pm 0.082$ & $0.540 \pm 0.199$ & $0.450 \pm 0.391$ & $0.243 \pm 0.145$ & $0.150 \pm 0.035$ \\
                                   & SSSNET+Topo & $0.872 \pm 0.125$ & $0.946 \pm 0.009$ & $0.936 \pm 0.081$ & $0.972 \pm 0.022$ & $0.388 \pm 0.164$ & $0.241 \pm 0.130$ \\

\midrule
\multirow{4}{*}{$\text{size}=1.0$} & DSGC+Topo     & $0.762 \pm 0.133$ & $0.777 \pm 0.160$ & $0.911 \pm 0.083$ & $0.989 \pm 0.009$ & $0.472 \pm 0.044$ & $0.156 \pm 0.024$ \\
                                   & SigMaNet+Topo & $0.169 \pm 0.015$ & $0.248 \pm 0.062$ & $0.284 \pm 0.034$ & $0.458 \pm 0.234$ & $0.174 \pm 0.042$ & $0.072 \pm 0.006$ \\
                                   & MSGNN+Topo    & $0.684 \pm 0.022$ & $0.559 \pm 0.105$ & $0.674 \pm 0.054$ & $0.974 \pm 0.022$ & $0.435 \pm 0.084$ & $0.170 \pm 0.013$ \\
                                   & SSSNET+Topo   & $0.923 \pm 0.030$ & $0.923 \pm 0.029$ & $0.937 \pm 0.047$ & $0.967 \pm 0.028$ & $0.481 \pm 0.034$ & $0.298 \pm 0.041$ \\
\midrule
\multirow{4}{*}{$\text{size}=0.1$} & DSGC+Topo     & $0.758 \pm 0.139$ & $0.680 \pm 0.193$ & $0.739 \pm 0.264$ & $0.991 \pm 0.016$ & $0.503 \pm 0.045$ & $0.169 \pm 0.036$ \\
                                   & SigMaNet+Topo & $0.136 \pm 0.090$ & $0.196 \pm 0.029$ & $0.300 \pm 0.065$ & $0.491 \pm 0.236$ & $0.092 \pm 0.084$ & $0.098 \pm 0.017$ \\
                                   & MSGNN+Topo    & $0.513 \pm 0.109$ & $0.474 \pm 0.114$ & $0.512 \pm 0.221$ & $0.915 \pm 0.133$ & $0.432 \pm 0.073$ & $0.170 \pm 0.014$ \\
                                   & SSSNET+Topo   & $0.888 \pm 0.137$ & $0.892 \pm 0.094$ & $0.724 \pm 0.304$ & $0.998 \pm 0.003$ & $0.442 \pm 0.060$ & $0.276 \pm 0.059$ \\
\bottomrule
\end{tabular}%
}
\end{table}
\vspace{-8mm}
\subsection{Ablation Study: Persistent Homology vs. Explicit Signed Degree Features}
\label{app_subsec:ablation_ph_vs_degree}
\vspace{-2mm}
To isolate the performance gain specifically contributed by persistent homology compared to non-topological degree signals, we construct an explicit signed-degree baseline. Specifically, for each node, we extract a 2-dimensional signed net degree descriptor consisting of the net in-degree and net out-degree, defined as $\text{Net In} = d_{in}^+ - d_{in}^-$ and $\text{Net Out} = d_{out}^+ - d_{out}^-$. To stabilize the degree distribution while preserving the directional sign of the net balance, we apply a signed logarithmic compression: $\text{sign}(x) \cdot \log(1 + |x|)$ for each dimension $x \in \{\text{Net In}, \text{Net Out}\}$. We then evaluate the model performance when using this explicit degree representation in place of the persistent homology features. Tables~\ref{tab:ablation_ph_vs_raw_degree_table1} and~\ref{tab:ablation_ph_vs_raw_degree_table2} present the downstream node clustering ARI across datasets under the prompt-based learning and semi-supervised signed GNN settings, respectively. This comparison verifies whether multi-scale homological summaries ($H_0$ connectivity and $H_1$ cycles) provide structural benefits beyond directional signed net degree statistics.
\begin{table}[htbp]
\centering
\caption{Pre-training and prompt learning performance comparison in terms of downstream node clustering ARI. We report the mean ARI among five runs plus/minus one standard deviation. Negative difference values in $\Delta$ are marked in \textbf{bold}.}
\label{tab:ablation_ph_vs_raw_degree_table1}
\resizebox{\textwidth}{!}{%
\begin{tabular}{llcccccc}
\toprule
\textbf{Features} & \textbf{Method} & \textbf{SDSBM-1} & \textbf{SDSBM-2} & \textbf{SDSBM-3} & \textbf{SDSBM-4} & \textbf{Rainfall} & \textbf{SP1500} \\
\midrule
\multirow{2}{*}{Degree-based} & TopoMSGNN & $0.026 \pm 0.031$ & $0.098 \pm 0.026$ & $0.370 \pm 0.041$ & $0.235 \pm 0.130$ & $0.073 \pm 0.060$ & $0.001 \pm 0.002$ \\
                              & TopoSSSNET & $0.039 \pm 0.019$ & $0.111 \pm 0.033$ & $0.381 \pm 0.019$ & $0.133 \pm 0.019$ & $0.045 \pm 0.055$ & $0.015 \pm 0.020$ \\
\midrule
\multirow{2}{*}{PH}           & \textbf{TopoMSGNN (Ours)} & $0.032 \pm 0.030$ & $0.107 \pm 0.009$ & $0.372 \pm 0.042$ & $0.164 \pm 0.025$ & $0.075 \pm 0.061$ & $0.000 \pm 0.000$ \\
                              & \textbf{TopoSSSNET (Ours)} & $0.076 \pm 0.031$ & $0.125 \pm 0.011$ & $0.379 \pm 0.036$ & $0.157 \pm 0.023$ & $0.062 \pm 0.057$ & $0.000 \pm 0.000$ \\
\midrule
\multirow{2}{*}{$\Delta$ (Degree $-$ PH)} & TopoMSGNN & $\mathbf{-0.006}$ & $\mathbf{-0.009}$ & $\mathbf{-0.002}$ & $+0.071$ & $\mathbf{-0.002}$ & $+0.001$ \\
                                         & TopoSSSNET & $\mathbf{-0.037}$ & $\mathbf{-0.014}$ & $+0.002$ & $\mathbf{-0.024}$ & $\mathbf{-0.017}$ & $+0.015$ \\
\bottomrule
\end{tabular}%
}
\end{table}
\vspace{-10mm}
\begin{table}[htbp]
\centering
\caption{Semi-supervised node clustering performance (ARI) on real-world and synthetic signed datasets. We compare our approach against state-of-the-art signed GNNs. We report the mean ARI among five runs plus/minus one standard deviation. The best method is marked in \textbf{bold} while the second best is marked with \underline{underline}.}
\label{tab:ablation_ph_vs_raw_degree_table2}
\resizebox{\textwidth}{!}{%
\begin{tabular}{llcccccc}
\toprule
\textbf{Features} & \textbf{Method} & \textbf{SDSBM-1} & \textbf{SDSBM-2} & \textbf{SDSBM-3} & \textbf{SDSBM-4} & \textbf{Rainfall} & \textbf{SP1500} \\
\midrule
\multirow{4}{*}{Degree-based} & DSGC+Topo     & $0.733 \pm 0.128$ & $0.760 \pm 0.227$ & $0.919 \pm 0.057$ & $\mathbf{0.989 \pm 0.009}$ & \underline{$0.491 \pm 0.058$} & $0.157 \pm 0.036$ \\
                              & SigMaNet+Topo & $0.145 \pm 0.090$ & $0.271 \pm 0.022$ & $0.310 \pm 0.066$ & $0.527 \pm 0.087$ & $0.146 \pm 0.145$ & $0.064 \pm 0.032$ \\
                              & MSGNN+Topo    & $0.677 \pm 0.013$ & $0.549 \pm 0.082$ & $0.695 \pm 0.035$ & \underline{$0.975 \pm 0.018$} & $0.433 \pm 0.085$ & $0.169 \pm 0.014$ \\
                              & SSSNET+Topo   & $\mathbf{0.948 \pm 0.015}$ & \underline{$0.858 \pm 0.125$} & \underline{$0.922 \pm 0.087$} & $0.971 \pm 0.021$ & $\mathbf{0.511 \pm 0.109}$ & \underline{$0.295 \pm 0.072$} \\
\midrule
\multirow{4}{*}{PH}           & DSGC+Topo     & $0.762 \pm 0.133$ & $0.777 \pm 0.160$ & $0.911 \pm 0.083$ & $\mathbf{0.989 \pm 0.009}$ & $0.472 \pm 0.044$ & $0.156 \pm 0.024$ \\
                              & SigMaNet+Topo & $0.169 \pm 0.015$ & $0.248 \pm 0.062$ & $0.284 \pm 0.034$ & $0.458 \pm 0.234$ & $0.174 \pm 0.042$ & $0.072 \pm 0.006$ \\
                              & MSGNN+Topo    & $0.684 \pm 0.022$ & $0.559 \pm 0.105$ & $0.674 \pm 0.054$ & $0.974 \pm 0.022$ & $0.435 \pm 0.084$ & $0.170 \pm 0.013$ \\
                              & SSSNET+Topo   & \underline{$0.923 \pm 0.030$} & $\mathbf{0.923 \pm 0.029}$ & $\mathbf{0.937 \pm 0.047}$ & $0.967 \pm 0.028$ & $0.481 \pm 0.034$ & $\mathbf{0.298 \pm 0.041}$ \\
\bottomrule
\end{tabular}%
}
\end{table}

\end{document}